\documentclass[letterpaper, 10 pt, conference]{ieeeconf}  

\IEEEoverridecommandlockouts                              

\usepackage[dvipsnames]{xcolor}
\usepackage{graphics} 
\usepackage{epsfig} 
\usepackage{amsmath} 
\usepackage{amssymb}  
\usepackage{bm}

\newcommand{\bb}{{\bm b}}

\newcommand{\bg}{{\bm g}}
\newcommand{\br}{{\bm r}}
\newcommand{\bu}{{\bm u}}

\newcommand{\bw}{{\bm w}}
\newcommand{\bx}{{\bm x}}

\newcommand{\bzeta}{{\bm \zeta}}
\newcommand{\beeta}{{\bm \eta}}
\newcommand{\btheta}{{\bm \theta}}
\newcommand{\blambda}{{\bm \lambda}}
\newcommand{\bmu}{{\bm \mu}}

\newcommand{\bSigma}{{\bm \Sigma}}

\newcommand{\bPsi}{{\bm \Psi}}

\newcommand{\vr}{{\bf r}}

\newcommand{\vG}{{\bf G}}

\newcommand{\vK}{{\bf K}}
\newcommand{\vL}{{\bf L}}

\newcommand{\vR}{{\bf R}}
\newcommand{\vS}{{\bf S}}
\newcommand{\vT}{{\bf T}}

\newcommand{\vW}{{\bf W}}

\usepackage{calrsfs}
\DeclareMathAlphabet{\pazocal}{OMS}{zplm}{m}{n}

\newcommand{\calC}{\pazocal{C}}

\newcommand{\calI}{\pazocal{I}}

\newcommand{\calL}{\pazocal{L}}

\newcommand{\calN}{\pazocal{N}}

\newcommand{\calO}{\pazocal{O}}

\newcommand{\calP}{\pazocal{P}}

\newcommand{\calU}{\pazocal{U}}

\newcommand{\calV}{\pazocal{V}}

\newcommand{\calW}{\pazocal{W}}

\newcommand{\calX}{\pazocal{X}}

\newcommand{\Eb}{\mathbb{E}}
\newcommand{\Pb}{\mathbb{P}}

\newcommand{\Rb}{\mathbb{R}}
\newcommand{\Sb}{\mathbb{S}}
\newcommand{\Zb}{\mathbb{Z}}

\newcommand{\vzero}{{\bf 0}}

\newcommand{\cov}{{\mathrm {Cov}}}
\newcommand{\trace}{{\mathrm {tr}}}
\newcommand{\diag}{{\mathrm {diag}}}
\newcommand{\bdiag}{{\mathrm {bdiag}}}
\newcommand{\T}{{\mathrm {T}}}

\newtheorem{proposition}{Proposition}
\newtheorem{remark}{Remark}
\newtheorem{problem}{Problem}

\usepackage{stmaryrd}

\newcommand{\argmin}{\operatornamewithlimits{argmin}}

\usepackage{algorithm}
\usepackage[noend]{algpseudocode}

\usepackage{eqparbox}

\usepackage{varwidth}

\usepackage{caption}

\usepackage{subcaption}

\usepackage{tikz}

\makeatletter
\let\NAT@parse\undefined
\makeatother
\usepackage[colorlinks=true, linkcolor=blue, citecolor=green, urlcolor=blue, 
            bookmarks=false, hypertexnames=true]{hyperref}
\hypersetup{
    colorlinks=blue,
    linkcolor=blue,
    filecolor=magenta,      
    urlcolor=blue,
    pdfpagemode=FullScreen,
    }

\title{\LARGE \bf
Distributed Model Predictive Covariance Steering
}

\author{Augustinos D. Saravanos$^{1}$, Isin M. Balci$^{2}$, Efstathios Bakolas$^{2}$ and Evangelos A. Theodorou$^{1}$ 
\thanks{$^{1}$
Daniel Guggenheim School of Aerospace Engineering, Georgia Institute of Technology, GA, USA
        {\tt\small \{asaravanos,evangelos.theodorou\}@gatech.edu}}%
\thanks{$^{2}$
Department of Aerospace Engineering and Engineering Mechanics, University of Texas at Austin, TX, USA
        {\tt\small \{isinmertbalci,bakolas@austin\}@utexas.edu, }}%
}

\begin{document}

\maketitle
\thispagestyle{empty}
\pagestyle{empty}

\begin{tikzpicture}[remember picture, overlay]
    \node[anchor=north west, yshift=-8mm, xshift=17mm] at (current page.north west)
    {\fontsize{8}{8}\selectfont Published at IEEE/RSJ International Conference on Intelligent Robots and Systems (IROS) 2024. Preprint version.};
\end{tikzpicture}

\begin{abstract}
This paper proposes Distributed Model Predictive Covariance Steering (DiMPCS) for multi-agent control under stochastic uncertainty. The scope of our approach is to blend covariance steering theory, distributed optimization and model predictive control (MPC) into a single framework that is safe, scalable and decentralized. Initially, we pose a problem formulation that uses the Wasserstein distance to steer the state distributions of a  multi-agent system to desired targets, and probabilistic constraints to ensure safety. We then transform this problem into a finite-dimensional optimization one by utilizing a disturbance feedback policy parametrization for covariance steering and a tractable approximation of the safety constraints. To solve the latter problem, we derive a decentralized consensus-based algorithm using the Alternating Direction Method of Multipliers. This method is then extended to a receding horizon form, which yields the proposed DiMPCS algorithm. Simulation experiments on a variety of multi-robot tasks with up to hundreds of robots demonstrate the effectiveness of DiMPCS. The superior scalability and performance of the proposed method is also highlighted through a comparison against related stochastic MPC approaches. Finally, hardware results on a multi-robot platform also verify the applicability of DiMPCS on real systems. 
A video with all results is \href{https://youtu.be/tzWqOzuj2kQ}{available}.
\end{abstract}

\section{INTRODUCTION}

Multi-robot control is a domain with a significant variety of applications such as swarm robotics \cite{panagou_multiagent}, multi-UAV navigation \cite{maza2009multi}, motion planning \cite{kantaros_multirobot}, underwater vehicles \cite{kyriakopoulos_underwater}, and so forth. As the scale and complexity of such systems continuously increases, some of the most desired attributes for algorithms designed to control these systems include \textit{safety under uncertainty}, \textit{scalability} and \textit{decentralization}.

Model predictive control (MPC) has found several successful multi-robot applications \cite{rey_lygeros2018fully, dai2017distributed, zhang2020improved}, thanks to its optimization-based nature and intrinsic feedback capabilities. 
In the case where stochastic disturbances are present, several stochastic MPC (SMPC) approaches have been proposed for handling them such as \cite{p:yan2021stochasticMPC, p:arcari2022stochasticMPC, p:schildbach2014scenarioSMPC, p:oldewurtel2008tractableSMPC}. Nevertheless, the literature in combining MPC with the steering of the state distribution of a system to exact targets for enhancing safety remains quite scarce \cite{p:okamoto2019smpc-cs, balci2022csMPPI, yin2022csMPPI}.

Covariance steering (CS) theory considers a class of stochastic optimal control problems, where the main objective is to steer the state mean and covariance of a system to desired targets. 
While initial CS approaches had dealt with infinite-horizon problems for linear time-invariant systems \cite{p:skelton1987covcontroltheory, p:skelton1992improvedcovariance}, finite-horizon CS methods that also address linear time-variant dynamics, have recently gained attention such as \cite{p:chen2015covariance1, kotsalis2021convex, p:balci2022exactcovariancewasserstein, liu2022optimal}. Several successful robotics applications of CS can be found in motion planning \cite{p:okamoto2019pathplanning}, trajectory optimization \cite{balci2022csMPPI, yin2022csMPPI}, multi-agent control \cite{saravanos2021distributed, Saravanos-RSS-23}, etc.

In SMPC based methods, it is typically the feed-forward control inputs that are treated as optimization variables, while the feedback gains are fixed to a stabilizing value for the closed-loop system \cite{p:arcari2022stochasticMPC}.
However, the state covariance cannot actively be steered with such methods, while fixed static feedback gains might perform poorly for time-varying dynamics.
Thus, control policies resulting from standard SMPC approaches might be suboptimal and/or overly conservative against safety criteria. On the contrary, CS methods yield the optimal feedback gains that steer the state covariance to the desired targets, thus providing more flexibility to satisfy optimality and safety guarantees at the same time. 

\begin{figure}[!t]
\centering
\vspace{0.5cm}
\setlength{\fboxrule}{1pt} 
\fbox{\includegraphics[width=0.46\textwidth, trim={9.5cm 2cm 8.5cm 2cm},clip]{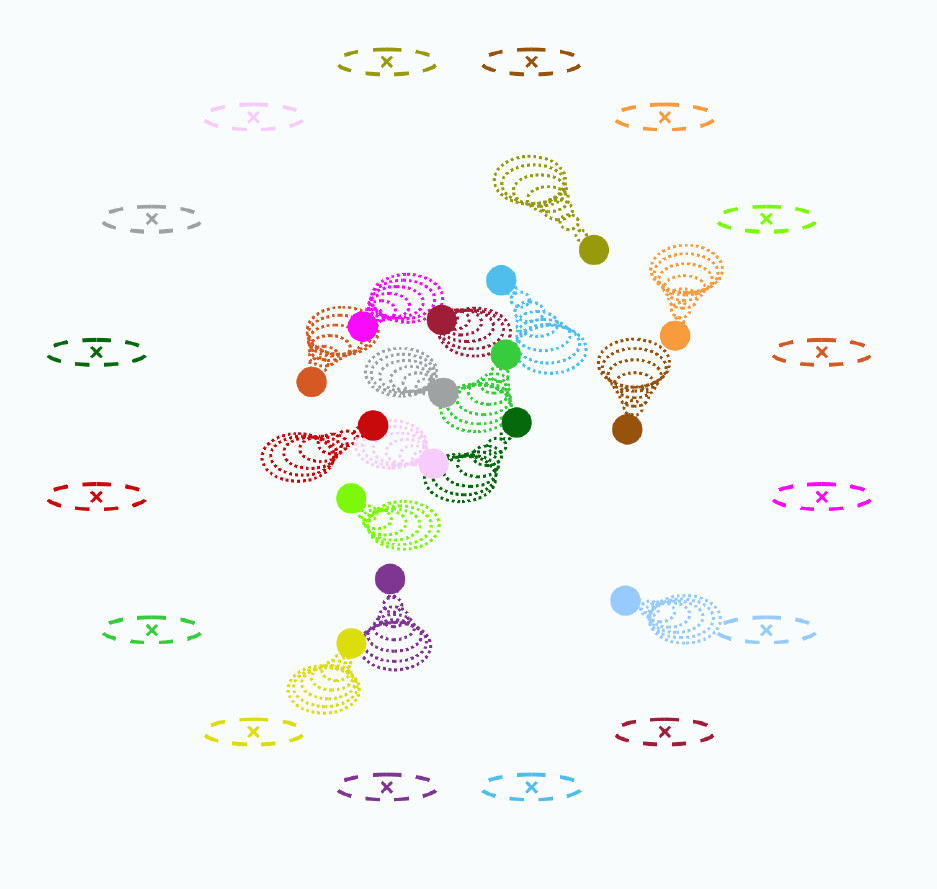}}
\caption{Sixteen unicycle robots safely guided with DiMPCS to their target distributions while avoiding collisions.}
\label{fig_intro}
\end{figure}


Although CS allows for finding the optimal control policies to steer the state statistics to desired values in the unconstrained case,  
the latter might be unreachable in the presence of state and/or input constraints. 
In MPC applications, especially, such infeasibilities can occur quite frequently, since the prediction horizon is usually much smaller than the total time horizon.
Therefore, it would be desirable to penalize the deviation from the desired state statistics by utilizing a distance metric between distributions such as the Wasserstein distance \cite{givens1984class}, instead of imposing hard constraints \cite{p:okamoto2019smpc-cs}. 

In addition, the main limitation of applying CS methods to large-scale multi-robot systems lies in the fact that computational demands increase significantly with respect to the state/control dimension and time horizon.
Nevertheless, recent work \cite{saravanos2021distributed} has shown that this computational burden can be significantly alleviated by merging CS with the Alternating Direction Method of Multipliers (ADMM), an optimization procedure that has found several recent applications in decentralized control \cite{halsted2021survey, cheng2021admm, saravanos2022distributed_ddp, pereira2022decentralized}.

In this paper, we propose Distributed Model Predictive Covariance Steering (DiMPCS) for safe and scalable multi-robot navigation. First, we provide a problem formulation which utilizes the Wasserstein distance for steering the robots to prescribed target distributions and probabilistic constraints for ensuring their safe operation. Subsequently, by exploiting CS theory, a suitable disturbance feedback policy parametrization, and an efficient approximation of the safety constraints, we transform the original problem into a finite-dimensional optimization one. To solve this, we propose an ADMM-based method for establishing consensus between neighboring robots and achieving decentralization. The latter method is then extended to an MPC scheme, which yields the final DiMPCS algorithm. Simulation experiments on several multi-agent navigation tasks with up to hundreds of robots illustrate the efficacy and scalability of DiMPCS. In addition, the advantages of the proposed method in terms of scalability and safety performance are also underlined through comparing with related SMPC approaches. Finally, we provide hardware experiments on a multi-robot platform which verify the effectiveness of DiMPCS on actual systems.
\vspace{-0.3cm}

\section{Problem Description}
\label{sec: problem formulation}

\subsection{Notation}
The space of $n\times n$ symmetric, positive semi-definite (definite) matrices is denoted with $\mathbb{S}_{n}^{+}$ ($\mathbb{S}_{n}^{++}$). 
The $n\times n$ identity matrix is denoted as $I_{n}$ whereas $\mathbf{0}$ denotes the zero matrix (or vector) with appropriate dimensions. 
The trace operator is denoted with $\trace (\cdot)$. 
The expectation and covariance of a random variable (r.v.) $x \in \Rb^n$ are given by $\Eb[x] \in \Rb^n$ and $\cov[x] \in \Sb^+_n$, respectively.
With $x \sim \calN (\mu, \Sigma) \in \Rb^n$, we refer to a Gaussian r.v. $x$ with $\Eb[x] = \mu$ and $\cov[x] = \Sigma$.
With $\llbracket a, b \rrbracket$, we denote the integer set $[a,b] \cap \Zb$ for any $a,b \in \Rb$.
The cardinality of a set $\calX$ is denoted with $|\calX|$. Finally, given a set $\calC$, we denote with $\calI_{\calC}(x)$ the indicator function such that $\calI_{\calC}(x) = 0$ if $x \in \calC$ and $\calI_{\calC}(x) = + \infty$, otherwise.

\subsection{Problem Description}

Let us consider a team of $N$ robots given by the set $\calV = \{ 1, \dots, N \}$. Each robot $i \in \calV$ is subject to the following  discrete-time, stochastic, nonlinear dynamics
%
%
%
\begin{equation}
x_{i,k+1} = f_i(x_{i,k}, u_{i,k}) + w_{i,k}, \quad 
x_{i,0} \sim \calN_{i,0},
\label{nonlinear dynamics}
\end{equation}
for $k \in \llbracket 0, K \rrbracket$, where $K$ is the time horizon, $x_{i,k} \in \Rb^{n_i}$, $u_{i,k} \in \Rb^{m_i}$ and $f_i: \Rb^{n_i \times m_i} \rightarrow \Rb^{n_i}$ are the state, control input and transition dynamics of the $i$-th robot, and $w_{i,k} \sim \calN(0, W_i)$ with $W \in \Sb_{n_i}^+$. Each robot's initial state $x_{i,0} \sim \calN_{i,0} = \calN(\mu_{i,0}, \Sigma_{i,0})$ with $\mu_{i,0} \in \Rb^{n_i}$ and $\Sigma_{i,0} \in \Sb_{n_i}^{+}$.

The position of the $i$-th robot in 2D (or 3D) space is denoted with $p_{i,k} \in \Rb^{q}$ with $q=2$ (or $q=3$) and can be extracted with $p_{i,k} = H_i x_{i,k}$, where $H_i \in \Rb^{q \times n_i}$ is defined accordingly. Furthermore, the environment, wherein the robots operate, includes circle (in 2D) or spherical (in 3D) obstacles given by the set $\calO = \{ 1, \dots, O \}$, where each obstacle $o \in \calO$ has position $p_o \in \Rb^{q}$ and radius $r_o \in \Rb$. 

We consider the problem of steering the state distributions of all robots $i \in \calV$ to the target Gaussian ones $\calN_{i,\mathrm{f}} = \calN (\mu_{i,\mathrm{f}}, \Sigma_{i, \mathrm{f}})$ with $\mu_{i,\mathrm{f}} \in \Rb^{n_i}$, $\Sigma_{i,\mathrm{f}} \in \Sb_{n_i}^{++}$. To penalize the deviation of the actual distributions from the target ones, we utilize the notion of the Wasserstein distance as a metric to describe similarity between r.v. probability distributions \cite{givens1984class}. In particular, we define the following cost:
\begin{equation}
J_{i} := 
\sum_{k=1}^{K} \calW_2^2(x_{i,k}, x_{i, \mathrm{f}}) + \Eb \Big[ \sum_{k=0}^{K-1} u_{i,k}^\T R_i u_{i,k} \Big],
\end{equation}
for each robot $i \in  \calV$, where $x_{i,\mathrm{f}} \sim \calN_{i,\mathrm{f}}$, $\calW_2^2(x_a, x_b)$ is the squared Wasserstein distance between $x_a, x_b$ and $R_i \in \Sb_{m_i}^{++}$.

The following probabilistic collision avoidance constraints between the robots and the obstacles are also imposed
\begin{align}
& \Pb ( \| p_{i,k} - p_o \|_2  \geq d_{i,o} + r_o ) \geq 1-\alpha , \nonumber \\
& \qquad \forall \ k \in \llbracket 0, K \rrbracket, \ i \in \calV, \ o \in \calO,
\label{obs avoidance}
\end{align}
where $0 < \alpha < 0.5$ and $d_{i,o} \in \Rb$ is the minimum allowed distance between the center of robot $i$ and obstacle $o$. In addition, we also wish for all robots to avoid collisions with each other, through the following constraints
\begin{align}
& \Pb ( \| p_{i,k} - p_{j,k} \|_2  \geq d_{i,j}) \geq 1-\alpha , \nonumber \\
& ~ \forall \ k \in \llbracket 0, K \rrbracket, \ i \in \calV, \ j \in \calV \backslash \{i\},
\label{collision avoidance}
\end{align}
where $d_{i,j} \in \Rb$ is the minimum allowed distance between the centers of the robots $i$ and $j$. 

Let us also define the sets of admissible control policies of the robots.
A control policy for robot $ i \in \calV$ is a sequence $\pi_i = \{ \tau_{i,0}, \tau_{i,1}, \dots, \tau_{i,K} \}$ where each $\tau_{i,k} : \mathbb{R}^{n_i(k + 1)} \rightarrow \mathbb{R}^{m_i}$ is a function of $\mathcal{X}_{i,0:k} = \{x_{i,0}, \dots, x_{i,k}\}$ that is the set of states already visited by robot $i$ at time $k$. 
The set of admissible policies for robot $i$ is denoted as $\Pi_i$. Finally, any additional control constraints we wish to impose are represented as $u_{i,k} \in \calU_i$. The multi-robot distribution steering problem can now be formulated as follows. 

\begin{problem}[Multi-Robot Distribution Steering Problem]
\label{general problem}
Find the optimal control policies $\pi_i^*, \ \forall i \in \calV$, such that 
\begin{align*}
    & ~~~~~~~~~~ \{ \pi_i^* \}_{i \in \calV} = \argmin \sum_{i\in \calV} J_i(\pi_{i}) \\
    \mathrm{s.t.} & ~~ \eqref{nonlinear dynamics}, \eqref{obs avoidance}, \eqref{collision avoidance}, ~ u_{i,k} = \tau_{i,k}(\mathcal{X}_{i}^{k}) \in \Pi_i, ~ u_{i,k} \in \calU_i, \ i \in \calV. \label{eq:control-input-constr}
\end{align*}
\end{problem}



\section{Multi-Agent Covariance Steering With Wasserstein Distance}

The scope of this work is to address Problem \ref{general problem} through leveraging CS theory, MPC and distributed optimization. While CS methods have mainly been developed for linear dynamics, they can be extended for nonlinear ones by linearizing around the mean of some reference trajectory \cite{p:ridderhof2019nonlinearcovariance, p:bakolas2020greedynonlinearcovariance, p:yi2020ddpcovariance}.
After linearization, 
we utilize a disturbance feedback policy parametrization which yields closed form expressions for the state means and covariances.
Finally, we transform Problem \ref{general problem} to an approximate finite-dimensional optimization one over the new policy parameters. 

\subsection{Dynamics Linearization}

By considering the first-order Taylor expansion of $f_i(x_{i,k}, u_{i,k})$ around some nominal trajectories $\bx_i' = [ x_{i,0}' ; \dots ; x_{i,K}' ], \ \bu_i' = [ u_{i,0}' ; \dots ; u_{i,K-1}' ]$, we obtain the discrete-time, stochastic, linear time-variant dynamics
%
%
\begin{equation}
x_{i,k+1} = A_{i,k} x_{i,k} + B_{i,k} u_{i,k} + r_{i,k} + w_{i,k}, ~~~ x_{i,0} \sim \calN_{i,0},
\end{equation}
where $A_{i,k} \in \Rb^{n_i \times n_i}$, $B_{i,k} \in \Rb^{n_i \times m_i}$ and $r_{i,k} \in \Rb^{n_i}$ are given by
%
%
\begin{subequations}
\begin{align}
A_{i,k} &= \left.\frac{\partial f}{\partial x_k} \right|_{\substack{x_k=x_k' \\ u_k=u_k'}, },
\quad
B_{i,k} = \left.\frac{\partial f}{\partial u_k} \right|_{\substack{x_k=x_k' \\ u_k=u_k'}, },
\label{linearization A B}
\\[0.1cm]
%
%
r_{i,k} &= f_i(x_{i,k}', u_{i,k}') - A_{i,k} x_{i,k}' - B_{i,k} u_{i,k}'.
\label{linearization residuals}
\end{align}
\end{subequations}
Therefore, each state trajectory can be expressed as 
\begin{equation}
\bx_i = \vG_{i,0} x_{i,0} + \vG_{i,u} \bu_i + \vG_{i,w} \bw_i + \vG_{i,w} \br_i,
\label{state compact}
\end{equation}
where $\bx_i = [x_{i,0}; \dots; x_{i,K}] \in \Rb^{(K+1)n_i}$, $\bu_i = [u_{i,0}; \dots; u_{i,K-1}] \in \Rb^{K m_i}$, $\bw_i = [w_{i,0}; \dots; w_{i,K-1}] \in \Rb^{K n_i}$, $\br_i = [r_{i,0}; \dots; r_{i,K-1}] \in \Rb^{K n_i}$, and the matrices 
$\mathbf{G}_{i,0} 
$, 
$\mathbf{G}_{i,u} 
$ and 
$
\mathbf{G}_{i,w} 
$ can be found in Eq. (9), (10) in \cite{balci2021letters}.

\subsection{Controller Parametrization}

Let us now consider the following affine disturbance feedback control policies, introduced in \cite{balci2021covariance}, 
%
\begin{equation}
u_{i,k} = \bar{u}_{i,k} + L_{i,k} (x_{i,0} - \mu_{i,0}) + \sum^{k-1}_{l=0} K_{i,(k-1,l)} w_{i,l},
\end{equation}
where $\bar{u}_{i,k} \in \Rb^{m_i}$ are the feed-forward parts of the control inputs and $L_{i,k}, \ K_{i,(k-1,l)} \in \Rb^{m_i \times n_i}$ are feedback matrices. 
Here, we assume perfect state measurements, such that the disturbances that have acted upon the system can be obtained.
It follows that
%
$
\bu_i = \bar{\bu}_i + \vL_i (x_{i,0} - \mu_{i,0}) + \vK_i \bw_i,
$
%
where $\bar{\bu}_i = [\bar{u}_{i,0}; \dots; \bar{u}_{i,K-1}] \in \Rb^{K m_i}$ and $\vL_i \in \Rb^{K m_i \times n_i}$, $\vK_i \in \Rb^{K m_i \times K n_i}$ are given by
%
%
%
$\vL_i = [L_{i,0}; \dots; L_{i,K-1}]$ and
\begin{equation*}
\vK_i = 
\begin{bmatrix}
\vzero & \vzero & \dots & \vzero & \vzero
\\
K_{i,(0,0)} & \vzero & \dots & \vzero & \vzero
\\
K_{i,(1,0)} & K_{i,(1,1)} & \dots & \vzero & \vzero
\\
\vdots & \vdots & \ddots & \vdots & \vdots
\\
K_{i,(K-2,0)} & K_{i,(K-2,1)} & \dots & K_{i,(K-2,K-2)} & \vzero
\end{bmatrix}.
\end{equation*}
Thus, the state trajectory of the $i$-th robot is obtained with
\begin{align}
\bx_i & = \vG_{i,0} x_{i,0} + \vG_{i,u} \bar{\bu}_i + \vG_{i,u} \vL_i (x_{i,0} - \mu_{i,0})
\nonumber
\\
& ~~~~~~~~~~~~~~~~~~ + (\vG_{i,w} + \vG_{i,u} \vK_i) \bw_i + \vG_{i,w} \vr_i.
\end{align}
Each state $x_{i,k}$ can be extracted with $x_{i,k} = \vT_{i,k} \bx_i$, where $\vT_{i,k} := \begin{bmatrix}
\vzero, \dots, I, \dots, \vzero
\end{bmatrix} \in \Rb^{n_i \times (K+1)n_i}$ is a block matrix whose $(k+1)$-th block is equal to the identity matrix and all the remaining blocks are equal to the zero matrix. Similarly, we also define $\vS_{i,k} \in \Rb^{m_i \times K m_i}$ such that $u_{i,k} = \vS_{i,k} \bu_i$.

\subsection{State Mean and Covariance Expressions}
Given that each state trajectory $\bx_i$ has been approximated as an affine expression of the Gaussian vectors $x_{i,0}$ and $\bw_i$, it follows that $\bx_i$ is also Gaussian, i.e., $\bx_i \in \calN(\bmu_i, \bSigma_i)$. With similar arguments as in \cite[Proposition 1]{balci2021covariance}, its mean $\bmu_i = \beeta_i(\bar{\bu}_i)$ and covariance $\bSigma_i = \btheta_i(\vL_i, \vK_i)$ are given by
%
%
%
\begin{align*}
\beeta_i(\bar{\bu}_i) & := \vG_{i,0} \mu_{i,0} + \vG_{i,u} \bar{\bu}_i + \vG_{i,w} \vr_i,
\\
\btheta_i(\vL_i, \vK_i) & := (\vG_{i,0} + \vG_{i,u} \vL_i) \Sigma_{i,0} (\vG_{i,0} + \vG_{i,u} \vL_i)^\T
\nonumber
\\
& ~~~~ + (\vG_{i,w} + \vG_{i,u} \vK_i) \vW_i (\vG_{i,w} + \vG_{i,u} \vK_i)^\T,
\end{align*}
where $\vW_i = \bdiag(W_i, \dots, W_i) \in \Rb^{K n_i \times K n_i}$. It follows that for each $x_{i,k} \sim \calN(\mu_{i,k}, \Sigma_{i,k})$, we have
%
$
\mu_{i,k} = \vT_{i,k} \beeta_i(\bar{\bu}_i)
$
and
$
\Sigma_{i,k} = \vT_{i,k} \btheta_i(\vL_i, ) \vT_{i,k}^\T.
$
%
It is important to note that the mean states depend only on the feed-forward control inputs $\bar{\bu}_i$, while the state covariances depend only on the feedback matrices $\vL_i, \vK_i$. 

\subsection{Problem Transformation}

%
%
%
%

The fact that the distributions of the states $x_{i,k}$ can be approximated as multivariate Gaussian ones, is of paramount importance here, since the Wasserstein distance admits a closed-form expression for Gaussian distributions - which does not hold for any arbitrary probability distributions \cite{givens1984class}. Therefore, we can rewrite each cost $J_i(\bar{\bu}_i, \vL_i, \vK_i)
= J_i^{\mathrm{dist}} (\bar{\bu}_i, \vL_i, \vK_i)
+ J_i^{\mathrm{cont}} (\bar{\bu}_i, \vL_i, \vK_i),$
%
%
where $J_i^{\mathrm{dist}}$ corresponds to the Wasserstein distances part and $J_i^{\mathrm{cont}}$ to the control effort part. Detailed expressions are provided in Appendix \ref{cost constraints appendix}.

Since the control input $u_{i,k}$ is a Gaussian r.v. as well, the control constraint $u_{i,k} \in \calU_i$ cannot be a hard constraint. 
For this reason, we use the following chance constraints instead,
\begin{align}
    \mathbb{P} ( \eta_{i,n}^\T u_{i,k} \leq \gamma_{i,n}) \geq 1 - \beta,
    \quad n = 1, \dots, N_u,
    \label{control chance constraint}
\end{align}
%
%
which yields the following convex quadratic constraint through the following proposition.

\begin{proposition}
The constraint \eqref{control chance constraint} can be equivalently expressed as
\begin{equation}
a_{i,n}(\bar{\bu}_i, \vL_i, \vK_i) \leq 0,
\end{equation}
with 
%
$
a_{i,n} = 
\eta_{i,n}^\T \vS_{i,k} \bar{\bu}_i - \gamma_{i,n} + \bar{\beta} \lVert \eta_{i,n}^\T \vS_{i,k} [\vL_i, \vK_i] \bPsi_i \rVert_2
$
%
and $\bPsi_i = \bdiag(\Sigma_{i,0}^{1/2}, \vW_i)$.
\end{proposition}
\begin{proof}
The proof is omitted as it follows similar steps as the one of \cite[Theorem 1]{okamoto2018optimal}.
\end{proof}

These constraints can be written more compactly for all $k \in \llbracket 0, K-1 \rrbracket$ as $a_i(\bar{\bu}_i, \vL_i, \vK_i) \leq 0$.

Finally, we also wish to express the collision avoidance constraints \eqref{obs avoidance}, \eqref{collision avoidance} w.r.t. the new decision variables. Starting from the obstacle avoidance ones, the chance constraint \eqref{obs avoidance} will always be satisfied if the following two constraints hold
%
%
\begin{align}
& \| \Eb[p_{i,k}] - p_o \|_2 \geq d_{i,o} + r_o, \quad \ i \in \calV, \ o \in \calO,
\label{obs avoidance 2 mean}
\\
& ~~~ d_{i,o} \geq 
\Bar{\alpha} \sqrt{ \lambda_{\mathrm{max}} \big(\bar{\Sigma}_{i,k} \big) },
\quad \ i \in \calV, \ o \in \calO,
\label{obs avoidance 2 covariance}
\end{align}
where $\bar{\Sigma}_{i,k} = H_i \Sigma_{i,k} H_i^\T$ is the position covariance, $\bar{\alpha} = \varphi^{-1}(\alpha)$ and $ \varphi^{-1}(\cdot)$ is the inverse of the cumulative density function of the normal distribution with unit variance. 
This is equivalent with enforcing that the $(\mu \pm \bar{\alpha} \sigma)$ confidence ellipsoid of the $i$-th robot's position is collision free.
In addition, since we are steering the covariances $\bar{\Sigma}_{i,k}$ to be as close as possible to the target $\bar{\Sigma}_{i,\mathrm{f}} = H_i \Sigma_{i,\mathrm{f}} H_i^\T$ through minimizing $J_i^{\mathrm{dist}} (\bar{\bu}_i, \vL_i, \vK_i)$, then assuming that the actual and target covariances will be close, we replace \eqref{obs avoidance 2 covariance} with 
%
\begin{equation}
d_{i,o} \geq 
\Bar{\alpha} \sqrt{ \lambda_{\mathrm{max}} \big(\bar{\Sigma}_{i,\mathrm{f}} \big) },
\quad \ i \in \calV, \ o \in \calO.
\label{obs avoidance 2 covariance b}
\end{equation}
Therefore, depending on the values of $\bar{\Sigma}_{i,\mathrm{f}}$ and $\bar{\alpha}$, we must choose a value for $d_{i,o}$ such that \eqref{obs avoidance 2 covariance b} will be satisfied, and then only the constraint \eqref{obs avoidance 2 mean} remains part of the optimization.


In a similar manner, the inter-robot collision avoidance chance constraints can be substituted with
\begin{equation}
\| \Eb[p_{i,k}] - \Eb[p_{j,k}] \|_2 \geq d_{i,j},~~\ i \in \calV, \ j \in \calV \backslash \{i\},
\label{collision avoidance 2 mean}
\end{equation}
\begin{equation}
d_{i,j} \geq 
\Bar{\alpha} \sqrt{ \lambda_{\mathrm{max}} \big(\bar{\Sigma}_{i,\mathrm{f}} \big) }
+ 
\Bar{\alpha} \sqrt{ \lambda_{\mathrm{max}} \big(\bar{\Sigma}_{j,\mathrm{f}} \big) },~~\ i \in \calV, \ j \in \calV \backslash \{i\}.
\label{collision avoidance 2 covariance}
\end{equation}
%
The constraints \eqref{obs avoidance 2 mean} and \eqref{collision avoidance 2 mean} can be written as $b_i(\bar{\bu}_i) \leq 0$ and $c_{i,j}(\bar{\bu}_i, \bar{\bu}_j) \leq 0$, respectively, with the exact expressions provided in Appendix \ref{cost constraints appendix}. Therefore, we arrive to the following tranformation of Problem \ref{general problem}.
\begin{problem}[Multi-Robot Distribution Steering Problem II]
\label{multi-robot problem 2}
Find the optimal feed-forward control sequences $\bar{\bu}_i^*$ and feedback matrices $\vL_i^*, \vK_i^*, \ \forall i \in \calV$, such that
\begin{subequations}
\begin{align}
& \{ \bar{\bu}_i^*, \vL_i^*, \vK_i^* \}_{i \in \calV} = \argmin \sum_{i \in \calV} J_i(\bar{\bu}_i, \vL_i, \vK_i)
\\
\mathrm{s.t.} \quad & a_i(\bar{\bu}_i, \vL_i, \vK_i) \leq 0, \quad b_i(\bar{\bu}_i) \leq 0, \quad \ i \in \calV,
\\
& c_{i,j}(\bar{\bu}_i, \bar{\bu}_j) \leq 0, \quad i \in \calV, \ j \in \calV \backslash \{i\}.
\label{collision avoidance compact}
\end{align}
\end{subequations}
\end{problem}


\section{Distributed Approach with ADMM}
\label{sec: distributed cs}

In this section, we present an ADMM-based methodology for solving Problem \ref{multi-robot problem 2} in a decentralized fashion. In this direction, we first introduce the notions of copy variables and consensus between neighboring robots, so that we can reformulate the problem in an equivalent form that is suitable for ADMM. Subsequently, the derivation of the ADMM updates is illustrated, yielding a distributed soft-constrained CS algorithm in a trajectory optimization format.  

\subsection{Decentralized Consensus Approach}
\label{subsec: distributed cs a}

Problem 1 cannot be solved directly in a distributed manner due to the inter-robot constraints \eqref{collision avoidance compact}. To address this issue, we first make the relaxation that each robot $i \in \calV$ only considers inter-robot constraints with its closest neighbors given by the set $\calV_i \subseteq \calV$ - defined such that $i \in \calV_i$ as well. Hence, the constraints \eqref{collision avoidance compact} can be replaced with 
%
$c_{i,j}(\bar{\bu}_i, \bar{\bu}_j) \leq 0, \ j \in \calV_i \backslash \{ i \}, \ i \in \calV$.
%
Subsequently, we introduce for each robot $i \in \calV$, the copy variables $\bar{\bu}_j^i$ regarding their neighbors $j \in \calV_i$. These copy variables can be interpreted as ``what is safe for robot $j$ from the perspective of robot $i$''. Thus, the augmented feed-forward control input $\bar{\bu}_i^{\mathrm{aug}} = [\{ \bar{\bu}_j^i \}_{j \in \calV_i}] \in \Rb^{K \tilde{m}_i}$ can be defined with $\tilde{m}_i = \sum_{j \in \calV_i} m_j$ . As a result, the inter-robot constraints can be rewritten from the perspective of the $i$-th robot as
%
$
c_{i,j}(\bar{\bu}_i, \bar{\bu}_j^i) \leq 0, ~ j \in \calV_i \backslash \{ i \}, \ i \in \calV,
$
%
or more compactly as $c_i^{\mathrm{aug}}(\bar{\bu}_i^{\mathrm{aug}}) \leq 0, ~ i \in \calV.$
%

Nevertheless, after the introduction of the copy variables, a requirement for enforcing a consensus between variables that refer to the same robot emerges. To accommodate this, let us define the global feed-forward control variable 
%
$
\bg = [\bg_1; \dots; \bg_N] \in \Rb^{Km}
$
%
where $m = \sum_{i \in \calV} m_i$. The necessary consensus constraints can be formulated as 
%
$
\bar{\bu}_j^i = \bg_j, \ j \in \calV_i \backslash \{i\}, \ i \in \calV,
$
%
or written more compactly as 
$
\bar{\bu}_i^{\mathrm{aug}} = \tilde{\bg}_i$, $i \in \calV,
%
$
where $\tilde{\bg}_i = [\{ \bg_j \}_{j \in \calV_i}] \in \Rb^{K \tilde{m}_i}$. Consequently, Problem \ref{multi-robot problem 2} can be rewritten in the following equivalent form. 
%
\begin{problem}[Multi-Robot Distribution Steering Problem III]
\label{multi-robot problem 3}
Find the optimal feed-forward control sequences $\bar{\bu}_i^{\mathrm{aug}*}$ and feedback matrices $\vL_i^*, \vK_i^*, \ \forall i \in \calV$, such that:
\begin{subequations}
\begin{align}
\{ \bar{\bu}_i^{\mathrm{aug}*}, & \vL_i^*, \vK_i^* \}_{i \in \calV} = \argmin \sum_{i \in \calV} J_i(\bar{\bu}_i, \vL_i, \vK_i)
\\
\mathrm{s.t.} \quad & a_i(\bar{\bu}_i, \vL_i, \vK_i) \leq 0,
\quad b_i(\bar{\bu}_i) \leq 0,  
\\
& c_i^{\mathrm{aug}}(\bar{\bu}_i^{\mathrm{aug}}) \leq 0, \quad \bar{\bu}_i^{\mathrm{aug}} = \tilde{\bg}_i, \quad i \in \calV.
\end{align}
\end{subequations}
\end{problem}
\vspace{0.2cm}
\begin{remark}
Since the inter-robot constraints only involve the feed-forward control inputs $\bar{\bu}_i$, then it is sufficient to add copy variables only for the latter - and not for $\vL_i, \vK_i$ as well. This is an important advantage of the policy parametrization we have selected, as in previous work \cite{saravanos2021distributed} where a state feedback parametrization was used, there was a requirement for consensus between both the feed-forward control inputs and the feedback gains, even in the case of mean inter-agent state constraints. Therefore, the affine disturbance feedback parametrization allows to significantly reduce the amount of optimization variables that each robot contains.
\end{remark}

\vspace{-0.1cm}
\subsection{Distributed Covariance Steering with Wasserstein Metric}
\label{subsec: distributed cs b}

Subsequently, let us proceed with the derivation of a decentralized ADMM algorithm for solving Problem \ref{multi-robot problem 3}. First, let us rewrite the problem in a more convenient form as
\begin{subequations}
\begin{align}
& \min \sum_{i \in \calV} J_i(\bar{\bu}_i, \vL_i, \vK_i) + \calI_{a_i}(\bar{\bu}_i, \vL_i, \vK_i) 
\nonumber
\\ & ~~~~~~~~~~~~~ + \calI_{b_i}(\bar{\bu}_i) 
+ \calI_{c_i^{\mathrm{aug}}}(\bar{\bu}_i^{\mathrm{aug}})
\\[0.2cm]
& ~~~~~~~~~~ \mathrm{s.t.} \quad 
\bar{\bu}_i^{\mathrm{aug}} = \tilde{\bg}_i, \quad i \in \calV.
\end{align}
\end{subequations}
The augmented Lagrangian (AL) is given by
\begin{align*}
\calL_{\rho} & = \sum_{i \in \calV} J_i(\bar{\bu}_i, \vL_i, \vK_i) + \calI_{a_i}(\bar{\bu}_i, \vL_i, \vK_i) 
+ \calI_{b_i}(\bar{\bu}_i) 
\nonumber
\\ & ~~~~~
+ \calI_{c_i^{\mathrm{aug}}}(\bar{\bu}_i^{\mathrm{aug}}) + \blambda_i^\T (\bar{\bu}_i^{\mathrm{aug}} - \tilde{\bg}_i)
+ \frac{\rho}{2} \| \bar{\bu}_i^{\mathrm{aug}} - \tilde{\bg}_i \|_2^2,
\end{align*}
where $\blambda_i$ are the dual variables for the constraints $\bar{\bu}_i^{\mathrm{aug}} = \tilde{\bg}_i$ and $\rho > 0$ is a penalty parameter.

\begin{algorithm*}[t]
\caption{Distributed Model Predictive Covariance Steering (DiMPCS)}\label{DMPCS Algorithm}
\begin{algorithmic}[1] 
\State \textbf{Set:} $N_{\mathrm{total}}$, $N_{\mathrm{pred}}$, $N_{\mathrm{comp}}, \ \rho, \ \ell_{\mathrm{max}}, \ \mu_{i,\mathrm{f}}, \ \Sigma_{i, \mathrm{f}}, \ R_i, \ \gamma_i, \ \forall i \in \calV$
\State $\hat{x}_{i,0} \leftarrow$ Measure initial robot states, $\forall i \in \calV$.
\State \textbf{Initialize:} 
$\ell \leftarrow 0$, $\bar{\bu}_{i|0}^{\mathrm{aug}} \leftarrow 0$, 
$\vL_{i|0} \leftarrow 0$, 
$\vK_{i|0} \leftarrow 0$,
$ \bg_{i|0} \leftarrow 0$, 
$ \blambda_{i|0} \leftarrow 0$,
$\bmu_{i|0} \leftarrow [ \hat{x}_{i,0}; \dots; \hat{x}_{i,0} ],
\ \forall i \in \calV$
\For{$k = 0, \dots, N_{\mathrm{total}}$}
\State $\hat{x}_{i,k} \leftarrow$ Measure current robot states, $\forall i \in \calV$.
\If{$\mathrm{mod}(k,N_{\mathrm{comp}}) == 0$}
%
\State $\calV_{i|k}, \calP_{i|k} \leftarrow$ Adapt neighborhoods based on current positions $\hat{p}_{i|k}$.
\Comment{In parallel $ \forall \ i \in \calV$}
\State
\begin{varwidth}[t]{\linewidth}
      $\{ A_{i,k'}, B_{i,k'}, r_{i,k'} \}_{k' \in \llbracket k, k + N_{\mathrm{pred}} -1 \rrbracket} \leftarrow $ Linearize dynamics using \eqref{linearization A B}, \eqref{linearization residuals}, around trajectories 
      $~~~~~~~~~~~~~~~~~~{\color{white}.}$
$\{ \mu_{i,k'}, \bar{u}_{i,k'} \}_{k' \in \llbracket k, k + N_{\mathrm{pred}} -1 \rrbracket}$. 
\Comment{In parallel $ \forall \ i \in \calV$}
      \end{varwidth}
\vspace{-0.1cm}
\State $\vG_{i,0|k}, \vG_{i,u|k}, \vG_{i,w|k} \leftarrow $ Construct using Eq. (9), (10) from \cite{balci2021letters}.
\Comment{In parallel $ \forall \ i \in \calV$}
\State $\mu_{i,k|k} \leftarrow \hat{x}_{i,k}, \ \Sigma_{i,k|k} \leftarrow 0, \ \ell \leftarrow 0$
\While{$\ell \leq \ell_{\mathrm{max}}$}
\State $\bar{\bu}_{i|k}^{\mathrm{aug}}, \vL_{i|k}, \vK_{i|k} \leftarrow$ Solve local optimization problem \eqref{local problem}.
\Comment{In parallel $ \forall \ i \in \calV$}
%
\State \textit{All robots  $j \in \calP_{i|k} \backslash \{i\}$ send $\bar{\bu}_{i|k}^j$ to each robot  $i \in \calV$.} 
\State $\bg_{i|k} \leftarrow$ Update with \eqref{global update}.
\Comment{In parallel $ \forall \ i \in \calV$}
\State \textit{All robots  $j \in \calV_{i|k} \backslash \{i\}$ send $\bg_{j|k}$ to each robot  $i \in \calV$.} 
\State $\blambda_{i|k} \leftarrow$ Update with \eqref{dual update}.
\Comment{In parallel $ \forall \ i \in \calV$}
\State $\ell \leftarrow \ell + 1$
\EndWhile
\State $\kappa \leftarrow k$
\EndIf
\State $u_{i,k} \leftarrow$ Compute with \eqref{mpc control policy} and apply control decision.
\Comment{In parallel $ \forall \ i \in \calV$}
\EndFor
\end{algorithmic}
\end{algorithm*}
%

In the first ADMM block, the AL is minimized w.r.t.  $\bar{\bu}_i^{\mathrm{aug}}$, $\vL_i$ and $\vK_i$, which yields the following $N$ local subproblems
%
%
%
\begin{align}
& \bar{\bu}_i^{\mathrm{aug}}, \vL_i, \vK_i \leftarrow 
\argmin_{\bar{\bu}_i^{\mathrm{aug}}, \vL_i, \vK_i} J_i(\bar{\bu}_i, \vL_i, \vK_i) + \blambda_i^\T (\bar{\bu}_i^{\mathrm{aug}} - \tilde{\bg}_i)
\nonumber
\\
& ~~~~~~~~~~~~~~~~~~~~~~~ + \frac{\rho}{2} \| \bar{\bu}_i^{\mathrm{aug}} - \tilde{\bg}_i \|_2^2 \label{local problem}
\\[0.2cm]
& \mathrm{s.t.} \quad  a_i(\bar{\bu}_i, \vL_i, \vK_i) \leq 0, \  b_i(\bar{\bu}_i) \leq 0, \ 
c_i^{\mathrm{aug}}(\bar{\bu}_i^{\mathrm{aug}}) \leq 0.
\nonumber
\end{align}
Note that each one of these subproblems can be solved in parallel by each robot $i$. Nevertheless, these are still non-convex problems due to the cost part $J_i^{\mathrm{dist}}$ and the constraints $b_i(\bar{\bu}_i) \leq 0$ and $c_i^{\mathrm{aug}}(\bar{\bu}_i^{\mathrm{aug}}) \leq 0$. In particular, as the cost $J_i^{\mathrm{dist}}$ is a sum of a convex and a concave term, we follow the same approach as in \cite{balci2021covariance} and solve the local problems with an iterative convex-concave procedure \cite{yuille2003concave}. In each such internal iteration, we also linearize the non-convex constraints around the previous mean trejectories as in \cite{augugliaro2012generation}. 


\begin{remark}
A significant advantage of using the squared Wasserstein distance as the measure of difference between actual and target distributions, is that 
the convexified version of \eqref{local problem}
is a convex quadratically constrained quadratic program (QCQP). 
This is in contrast with other CS approaches that yield semi-definite programs \cite{balci2021covariance, p:balci2022exactcovariancewasserstein, kotsalis2021convex} which are more computationally demanding to solve.
\end{remark}

In the second ADMM block, the AL is minimized w.r.t. $\bg$, which gives the ``per-robot'' update rules
\begin{equation}
\bg_i \leftarrow \frac{1}{|\calP_i|} \sum_{j \in \calP_i} \bar{\bu}_i^j + \frac{1}{\rho} \blambda_i^j,
\label{global update}
\end{equation}
where $\calP_i = \{ j \in \calV : i \in \calV_j \}$ defines the set that contains all robots $j \in \calV$ that have $i$ as a neighbor, and $\blambda_i^j$ is the part of the dual variable $\blambda_i$ that corresponds to the constraint $\bar{\bu}_j^i = \bg_j$. Finally, the dual variables are updated as follows
\begin{equation}
\blambda_i \leftarrow \blambda_i + \rho(\bar{\bu}_i^{\mathrm{aug}} - \tilde{\bg}_i),
\label{dual update}
\end{equation}
by all $i \in \calV$. The updates \eqref{local problem}, \eqref{global update} and \eqref{dual update} are repeated in the presented order until we reach to $\ell_{\mathrm{max}}$ iterations. 

%

\section{Distributed Model Predictive \\ Covariance Steering}
\label{sec: distributed mpc cs}

This section presents Distributed Model Predictive Covariance Steering (DiMPCS) which uses the method proposed in Section \ref{sec: distributed cs} at its core, by extending it in a receding horizon fashion. The full algorithm is presented in Algorithm \ref{DMPCS Algorithm}. 

Let us denote with $N_{\mathrm{total}}$ and $N_{\mathrm{pred}}$, the total and prediction time horizons, respectively. With $N_{\mathrm{comp}} \ (\leq N_{\mathrm{pred}})$, we set how often a new MPC computation is performed.
After setting all parameters (Line 1) and measuring the initial states $\hat{x}_{i,0}$ (Line 2), we initialize all decision variables with zeros, and the mean state trajectories with $\bmu_{i|0} \leftarrow [ \hat{x}_{i,0}; \dots; \hat{x}_{i,0} ]$ (Line 3). With the notation $z_{\cdot|k}$ we refer to any quantity $z$ that is computed at time $k$. 

Then, the control procedure starts for $k = 0, \dots, N_{\mathrm{total}}$. After measuring the current state if $k>0$ (Line 5), a new MPC computation starts if $\mathrm{mod}(k,N_{\mathrm{comp}}) = 0$. In this case, the neighborhood sets of all robots $i \in \calV$ are first found, by identifying the ones that are in close distance, based on their current positions (Line 7). Subsequently, the dynamics linearizaton (Line 8)
and the construction of the matrices $\vG_{i,0|k}, \vG_{i,u|k}, \vG_{i,w|k}$ (Line 9) take place. The mean $\mu_{i,k|k}$ is always initialized being equal with $\hat{x}_{i,k}$, while the initial covariance is set to $\Sigma_{i,k|k} = 0$ (Line 10) since in this MPC format, we have perfect information of the initial state $x_{i,k}$ before the optimization starts.


\begin{figure*}[!t]
     \centering
     \begin{subfigure}[b]{0.19\textwidth}
         \centering 
         \begin{tikzpicture}
        \node[anchor=south west,inner sep=0] at (0,0){    \includegraphics[width=\textwidth, trim={0cm 0cm 0cm 0cm},clip]{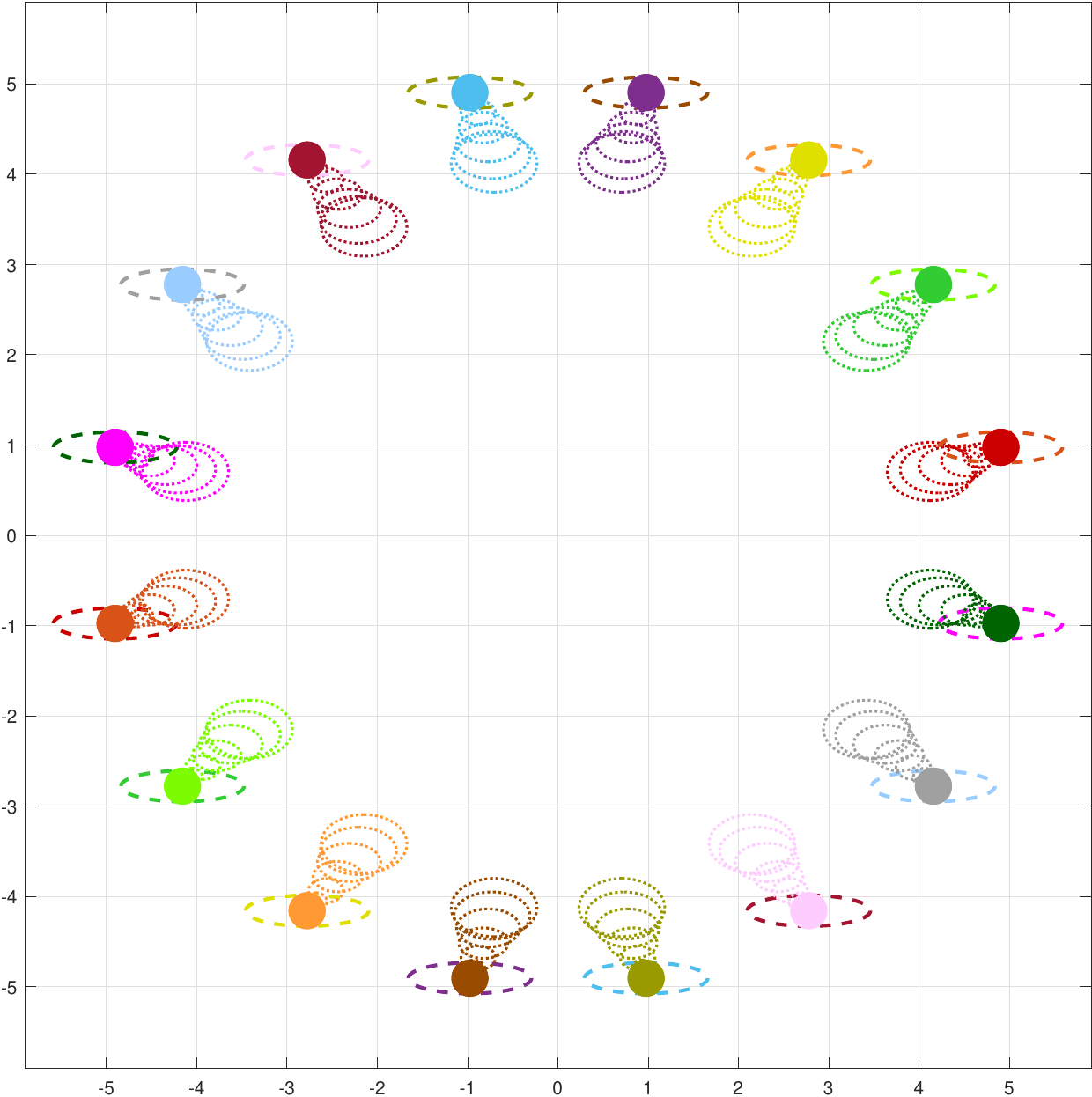}};
        \node[align=center, text=NavyBlue] (c) at (2.9, 0.3) {\small{$k = 0$}};
        \end{tikzpicture}
        \label{fig_swap_1}
        \end{subfigure}
     \begin{subfigure}[b]{0.19\textwidth}
         \centering 
          \begin{tikzpicture}
        \node[anchor=south west,inner sep=0] at (0,0){    \includegraphics[width=\textwidth, trim={0cm 0cm 0cm 0cm},clip]{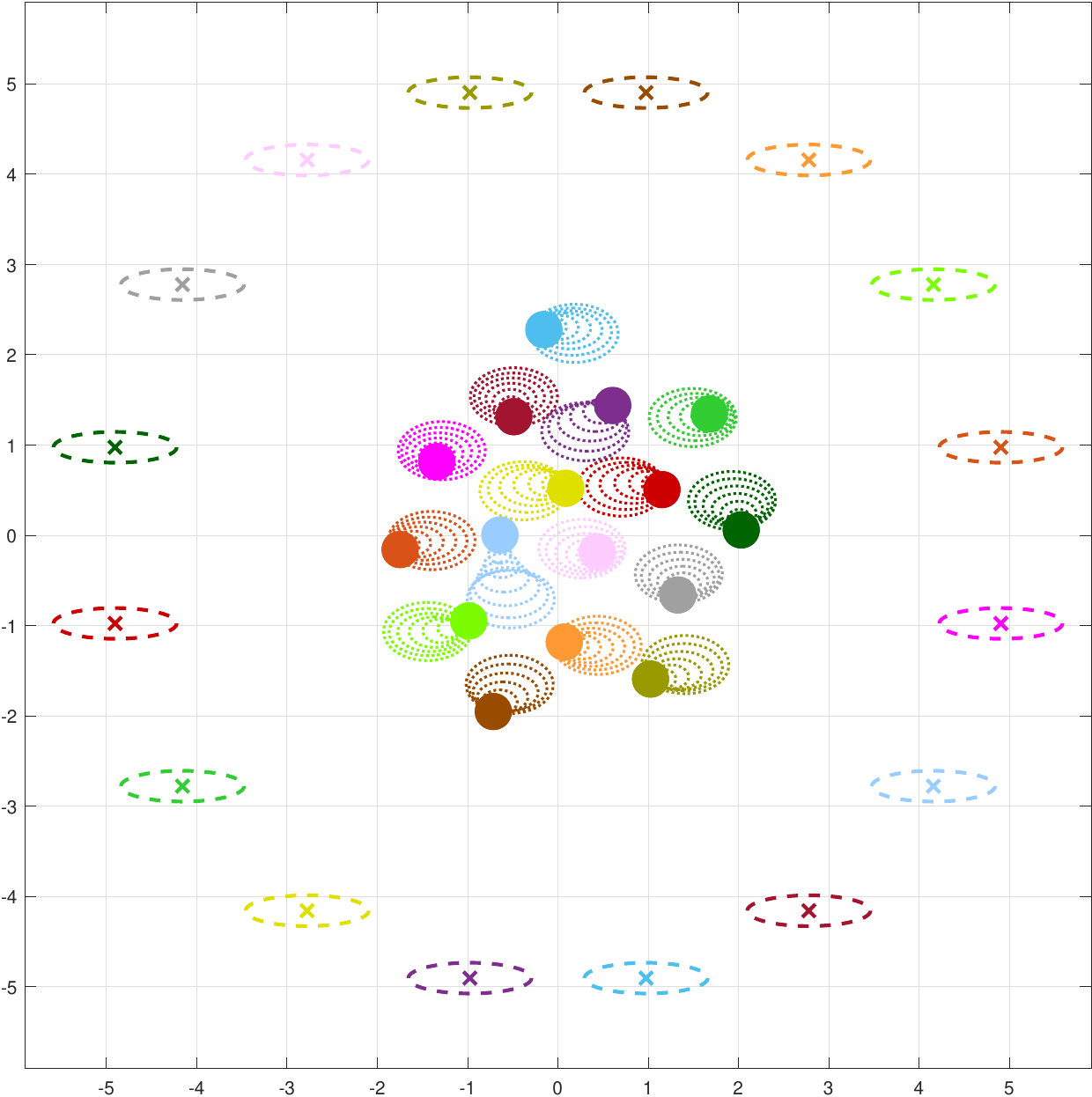}};
        \node[align=center, text=NavyBlue] (c) at (2.8, 0.3) {\small{$k = 20$}};
        \end{tikzpicture}
        \label{fig_swap_2}
    \end{subfigure}
     \centering
     \begin{subfigure}[b]{0.19\textwidth}
         \centering 
          \begin{tikzpicture}
        \node[anchor=south west,inner sep=0] at (0,0){    \includegraphics[width=\textwidth, trim={0cm 0cm 0cm 0cm},clip]{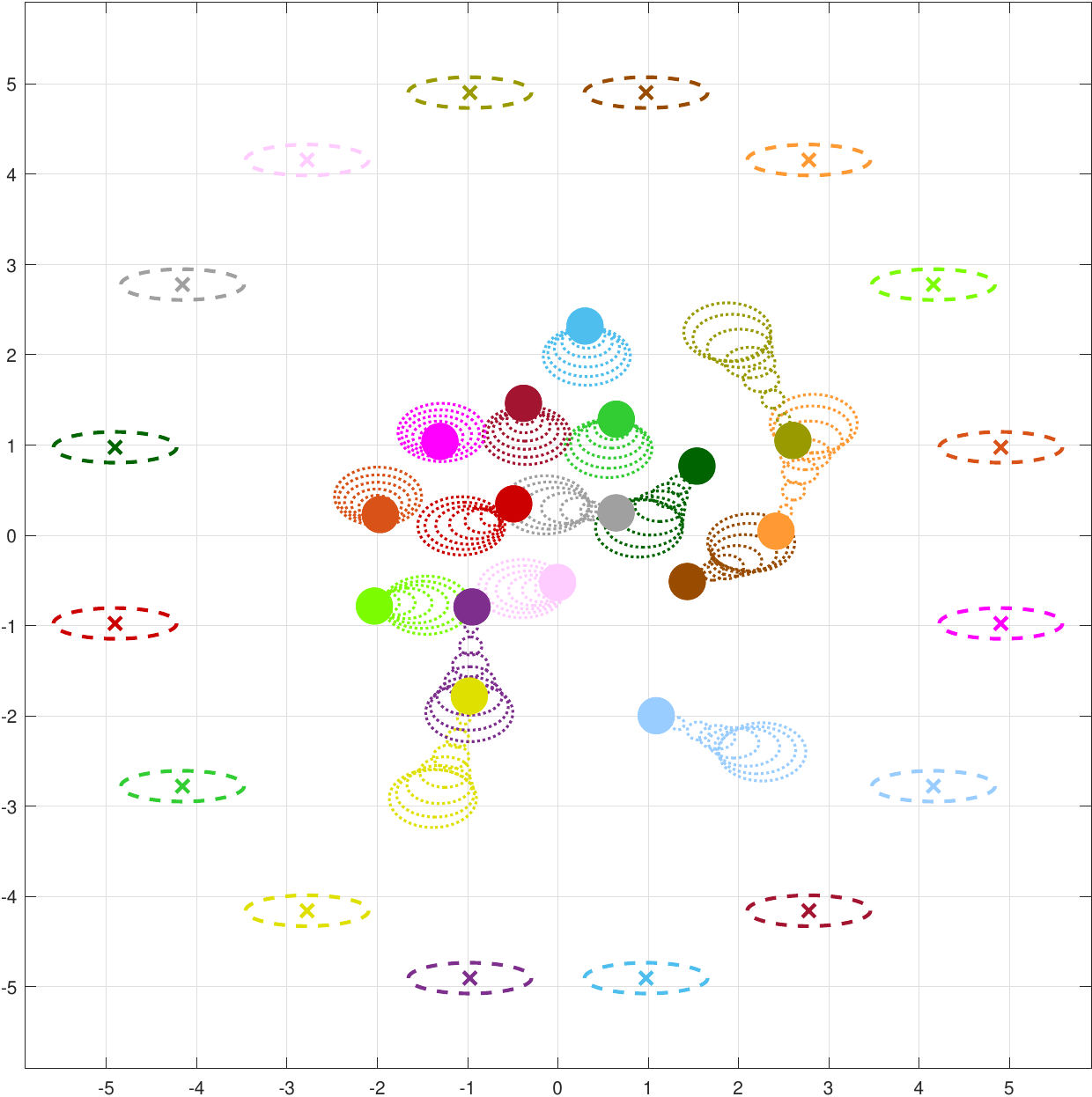}};
        \node[align=center, text=NavyBlue] (c) at (2.8, 0.3) {\small{$k = 40$}};
        \end{tikzpicture}
        \label{fig_swap_3}
    \end{subfigure}
     \begin{subfigure}[b]{0.19\textwidth}
         \centering 
          \begin{tikzpicture}
        \node[anchor=south west,inner sep=0] at (0,0){    \includegraphics[width=\textwidth, trim={0cm 0cm 0cm 0cm},clip]{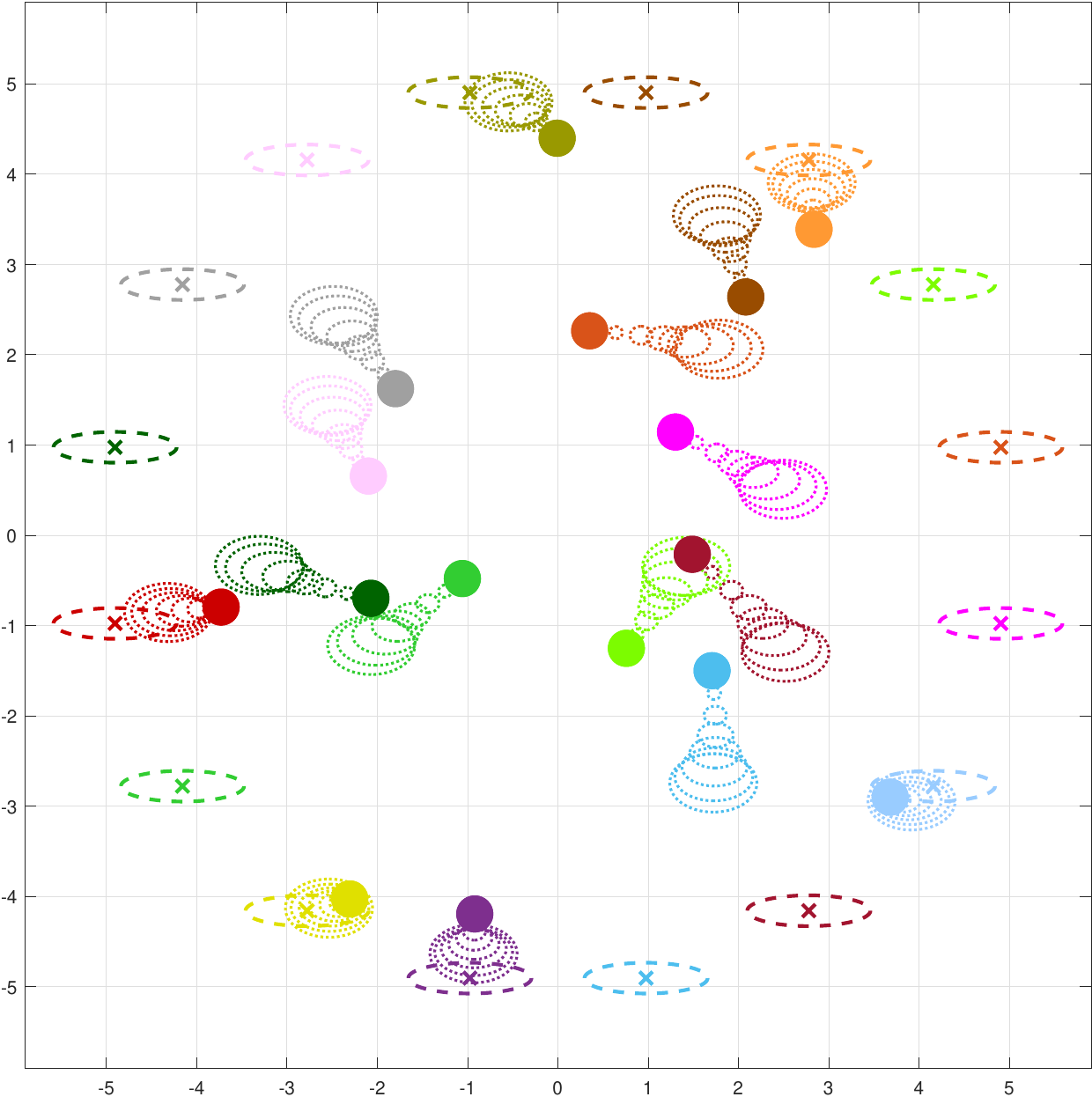}};
        \node[align=center, text=NavyBlue] (c) at (2.8, 0.3) {\small{$k = 60$}};
        \end{tikzpicture}
        \label{fig_swap_4}
    \end{subfigure}
     \begin{subfigure}[b]{0.19\textwidth}
         \centering 
          \begin{tikzpicture}
        \node[anchor=south west,inner sep=0] at (0,0){    \includegraphics[width=\textwidth, trim={0cm 0cm 0cm 0cm},clip]{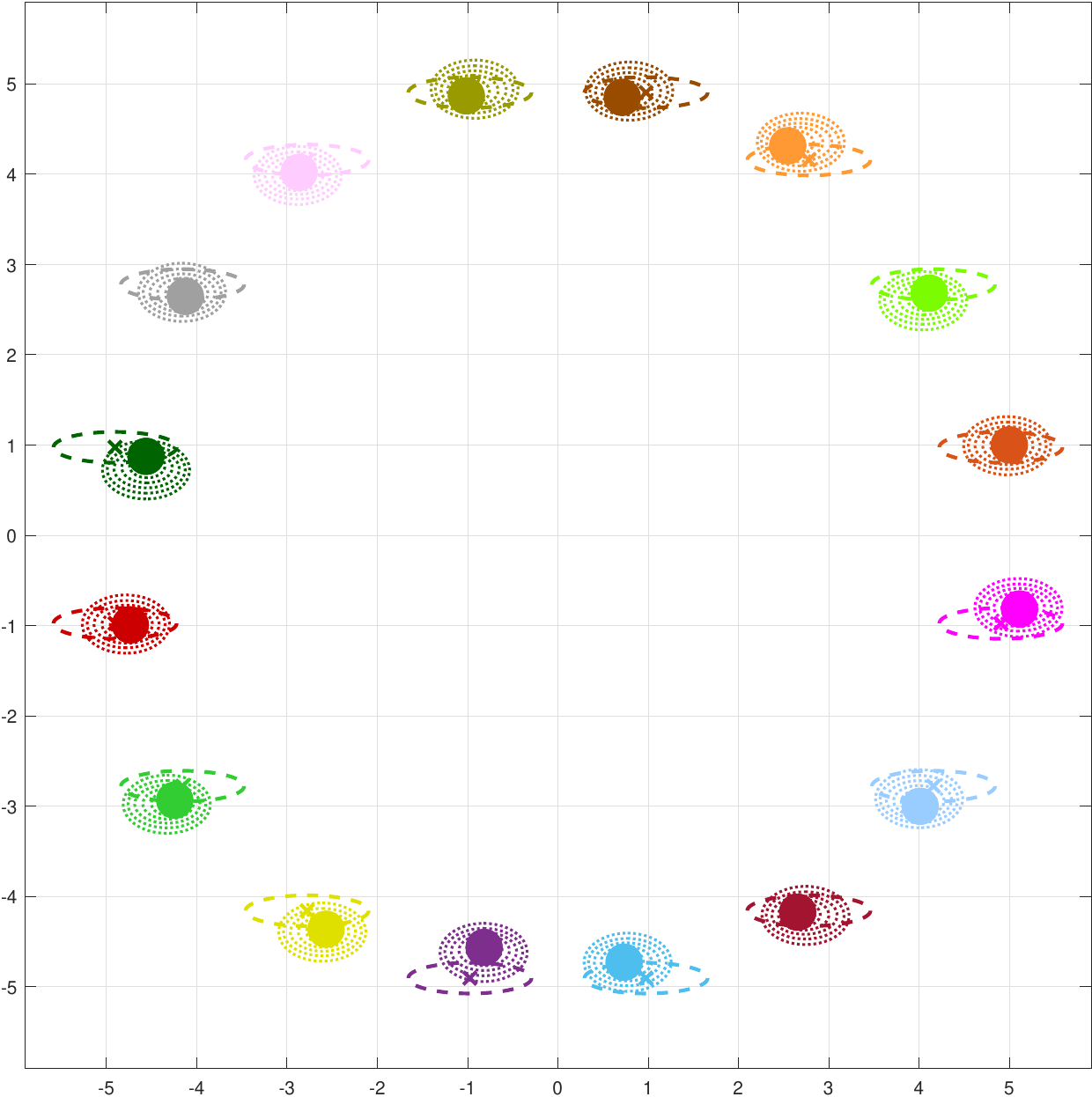}};
        \node[align=center, text=NavyBlue] (c) at (2.72, 0.3) {\small{$k = 150$}};
        \end{tikzpicture}
        \label{fig_swap_5}
    \end{subfigure}
    \vspace{-0.3cm}
\caption{A task with $16$ robots that must reach a target distribution at a diametrically opposite location while avoiding collisions. Each snapshot shows their positions and the $(\mu \pm 3 \sigma)$ confidence regions of planned distribution trajectories. The target distributions are shown as dashed ellipses with ``x'' at the center.}
\label{fig_sim_swap}
\end{figure*}

\begin{figure*}[!t]
     \centering
     \begin{subfigure}[b]{0.32\textwidth}
         \centering 
         \begin{tikzpicture}
        \node[anchor=south west,inner sep=0] at (0,0){    \includegraphics[width=\textwidth, trim={6.0cm 4.0cm 5cm 3cm},clip]{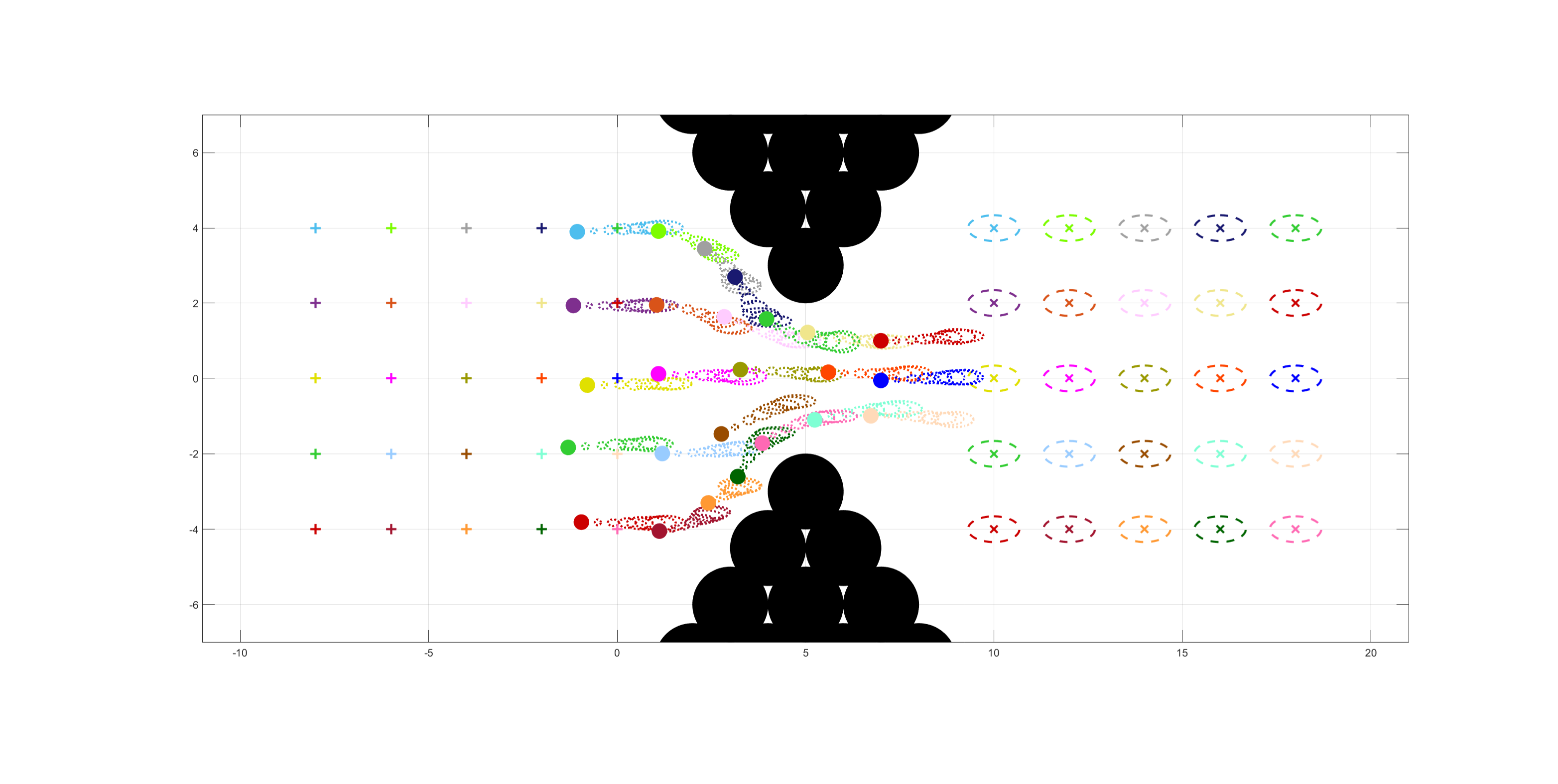}};
        \node[align=center, text=NavyBlue] (c) at (5.1, 0.32) {\small{$k = 20$}};
        \end{tikzpicture}
        \label{fig_bottle_1}
    \end{subfigure}
     \begin{subfigure}[b]{0.32\textwidth}
         \centering 
         \begin{tikzpicture}
        \node[anchor=south west,inner sep=0] at (0,0){    \includegraphics[width=\textwidth, trim={6.0cm 4.0cm 5cm 3cm},clip]{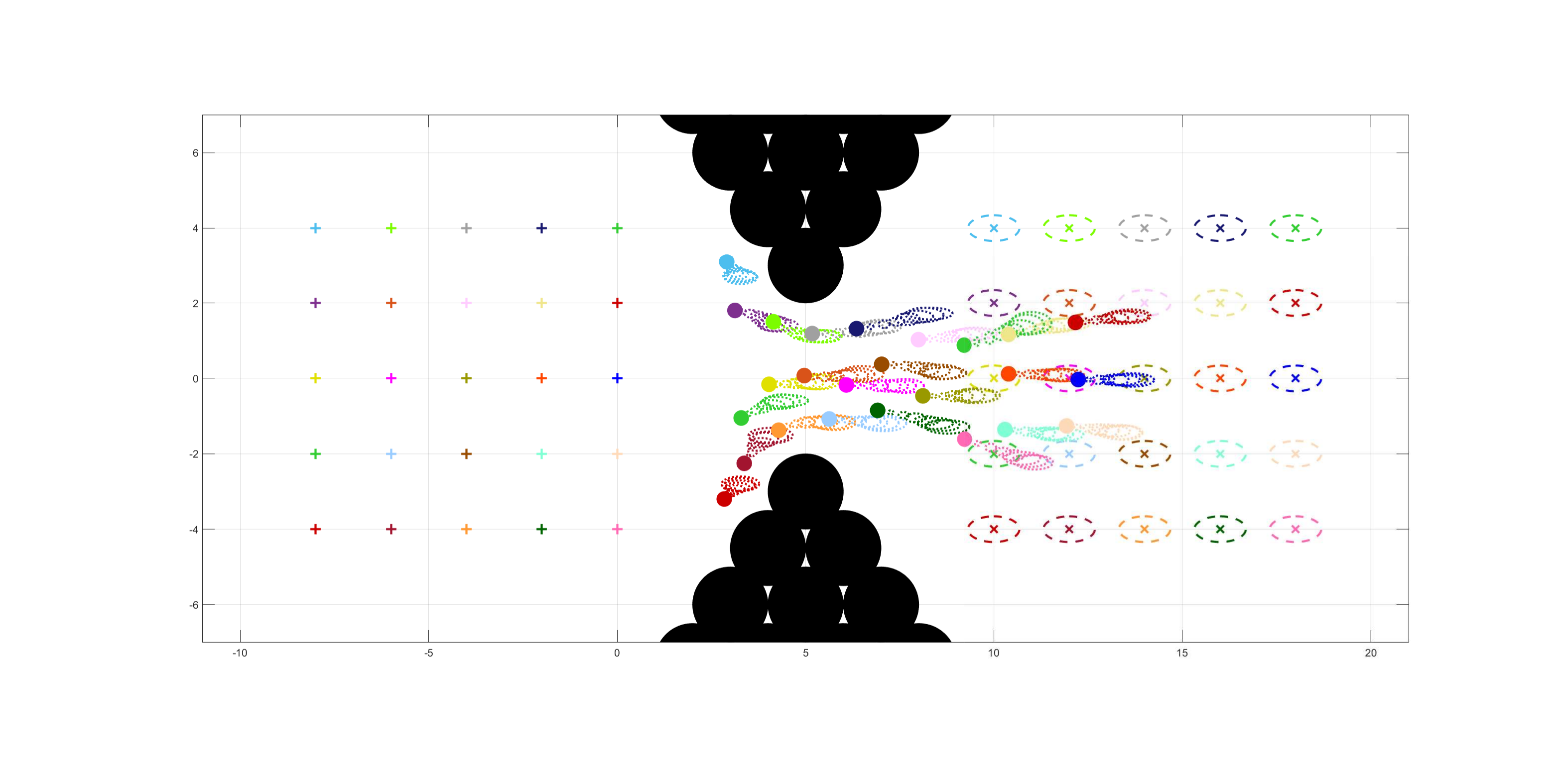}};
        \node[align=center, text=NavyBlue] (c) at (5.1, 0.32) {\small{$k = 40$}};
        \end{tikzpicture}
        \label{fig_bottle_2}
    \end{subfigure}
     \centering
     \begin{subfigure}[b]{0.32\textwidth}
         \centering 
         \begin{tikzpicture}
        \node[anchor=south west,inner sep=0] at (0,0){    \includegraphics[width=\textwidth, trim={6.0cm 4.0cm 5cm 3cm},clip]{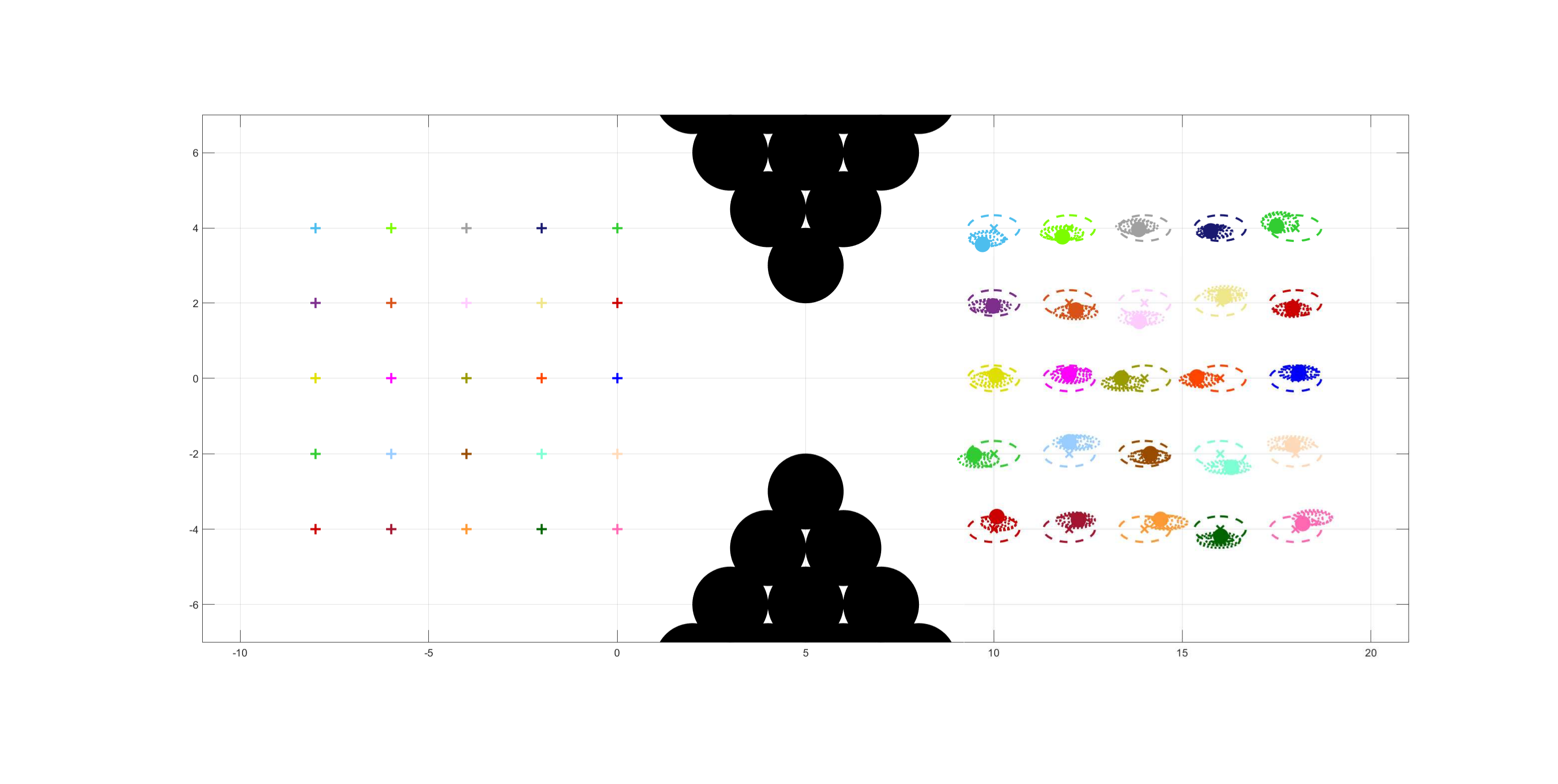}};
        \node[align=center, text=NavyBlue] (c) at (5.1, 0.32) {\small{$k = 90$}};
        \end{tikzpicture}
        \label{fig_bottle_3}
    \end{subfigure}
    \vspace{-0.3cm}
\caption{A task with $25$ robots required to pass through a narrow bottleneck before reaching their targets.}
\label{fig_sim_bottle}
\end{figure*}

The execution of the proposed ADMM method of Section \ref{sec: distributed cs} follows. First, the local decision variables $\bar{\bu}_{i|k}^{\mathrm{aug}}, \vL_{i|k}, \vK_{i|k}$ of each robot are obtained (Line 12) through solving the local CS problems \eqref{local problem} as explained in Section \ref{subsec: distributed cs b}. Afterwards, each robot $i$ receives the copy variables $\bar{\bu}_{i|k}^j$ from all $j \in \calP_{i|k} \backslash \{i\}$ (Line 13), so that it can compute $\bg_{i|k}$ (Line 14) with \eqref{global update}. Subsequently, each robot $i$ receives the variables $\bg_{j|k}$ from all $j \in \calV_{i|k} \backslash \{i\}$ (Line 15), so that  $\tilde{\bg}_{i|k}$ is constructed and the dual updates \eqref{dual update} take place (Line 16). This iterative ADMM procedure is terminated after $\ell_{\mathrm{max}}$ iterations. Finally, the control input of each robot is computed (Line 19) through
\begin{align}
u_{i,k|\kappa} &= \bar{u}_{i,k|\kappa} + L_{i,k|\kappa} (\hat{x}_{i,\kappa} - \mu_{i,\kappa|\kappa})
\nonumber
\\
& ~~~~~~~~~~~~~~~~~~~~~~~~~~ + \sum^{k-1}_{l=0} K_{i,(k-1,l)|\kappa} \ w_{i,l} 
\label{mpc control policy}
\end{align}
where $\kappa$ is the last time $k$ that an MPC cycle took place. Note that in the special case where we assign $\hat{x}_{i,\kappa} = \mu_{i,\kappa|\kappa}$, the second term in the RHS of \eqref{mpc control policy} becomes zero but this can change if the assumption of $\Sigma_{i,k|k} = 0$ is relaxed.

\begin{remark}
All computations in DiMPCS (Lines 7-9,12,14,16,19) can be performed in parallel by every robot $i \in \calV$. In addition, all necessary communication steps (Lines 13,15) take place locally between neighboring robots. Therefore, the proposed algorithm is \textit{fully distributed} in terms of computational and communication requirements.
\end{remark}
\begin{remark}
The neighborhood adaptation, during the beginning of every MPC cycle, is an important advantage compared to the trajectory optimization approach followed in \cite{saravanos2021distributed}, as it allows for using smaller adjustable neighborhoods. 
\end{remark}
\vspace{-0.09cm}




\section{Simulation Experiments}
\label{sec: simulations}

This section presents simulation experiments that demonstrate the effectiveness and scalability of DiMPCS. In the main paper, we provide snapshots of the tasks, while we refer the reader to the \href{https://youtu.be/tzWqOzuj2kQ}{supplementary video} for a full demonstration. All robots have unicycle dynamics with states $x_{i,k} = [ \mathrm{x}_{i,k}; \mathrm{y}_{i,k}; \theta_{i,k}; v_{i,k}]
\in \Rb^4$ 
and inputs 
$\bu_{i,k} = [
a_{i,k}; \omega_{i,k}]
\in \Rb^2$, 
where $(\mathrm{x}_{i,k}, \mathrm{y}_{i,k})$, $\theta_{i,k}$ $v_{i,k}$, $\omega_{i,k}$, $a_{i,k}$ are their 2D position coordinates, angles, linear and angular velocities and linear accelerations, respectively. In all experiments, we use $N_{\mathrm{pred}} = 7$ and $N_{\mathrm{comp}} = 2$. The discretization time step is $dt = 0.05$. The process noise covariance is $W_i = \diag(0.02,0.02,\pi/180,0.2)$. We set the control cost matrix $R_i = \diag(10^{-2}, 10^{-2})$. We also enforce control limits $a_{\text{max}} = - a_{\text{min}} = 5 \text{m/s}^2$ and $\omega_{\text{max}} = - \omega_{\text{min}} = 4 ~\text{rad/s}$ through the chance constraints \eqref{control chance constraint} with $\beta = 0.997$. For the collision avoidance constraints, we select $d_{i,o} = 0.75\text{m}$, $d_{i,j} = 1.5\text{m}$ and $\bar{\alpha} = \varphi^{-1}(0.997) = 3$. Finally, we set $\rho = 10^{-2}, \ \ell_{\mathrm{max}} = 30$ and $|\calV_i| = 6$ for all tasks. 

\subsection{Small-Scale Tasks}

In the first task, 16 robots need to reach to their target distributions at the diametrically opposite locations, while avoiding collisions with each other. In Fig. \ref{fig_sim_swap}, the performance of DiMPCS is demonstrated through five different snapshots that show the positions and planned distribution trajectories of the robots. All robots are able to successfully reach to their targets and avoid collisions throughout the task. In the next scenario (Fig. \ref{fig_sim_bottle}), 25 robots must reach to their targets while passing through a narrow ``bottleneck'' and avoiding collisions. Despite the difficulty of this task, all robots are again safely navigated to their targets.

\subsection{Large-Scale Task}

Subsequently, we highlight the scalability of DiMPCS to large-scale multi-robot problems. In particular, we consider a problem with 256 robots that need to move from one $16 \times 16$ square grid to another one while avoiding collisions with each other and the obstacles in between. Figure \ref{fig_formation} shows a snapshot of the task, while the full task is available in the supplementary video. All robots are successfully driven to their targets while maintaining their safe operation.

\subsection{Comparison with Other Stochastic MPC Approaches}

\begin{figure}[!t]
\centering
\begin{tikzpicture}
\node[anchor=south west,inner sep=0] at (0,0){    \includegraphics[width=0.49\textwidth, trim={6.1cm 3.0cm 5cm 3cm},clip]{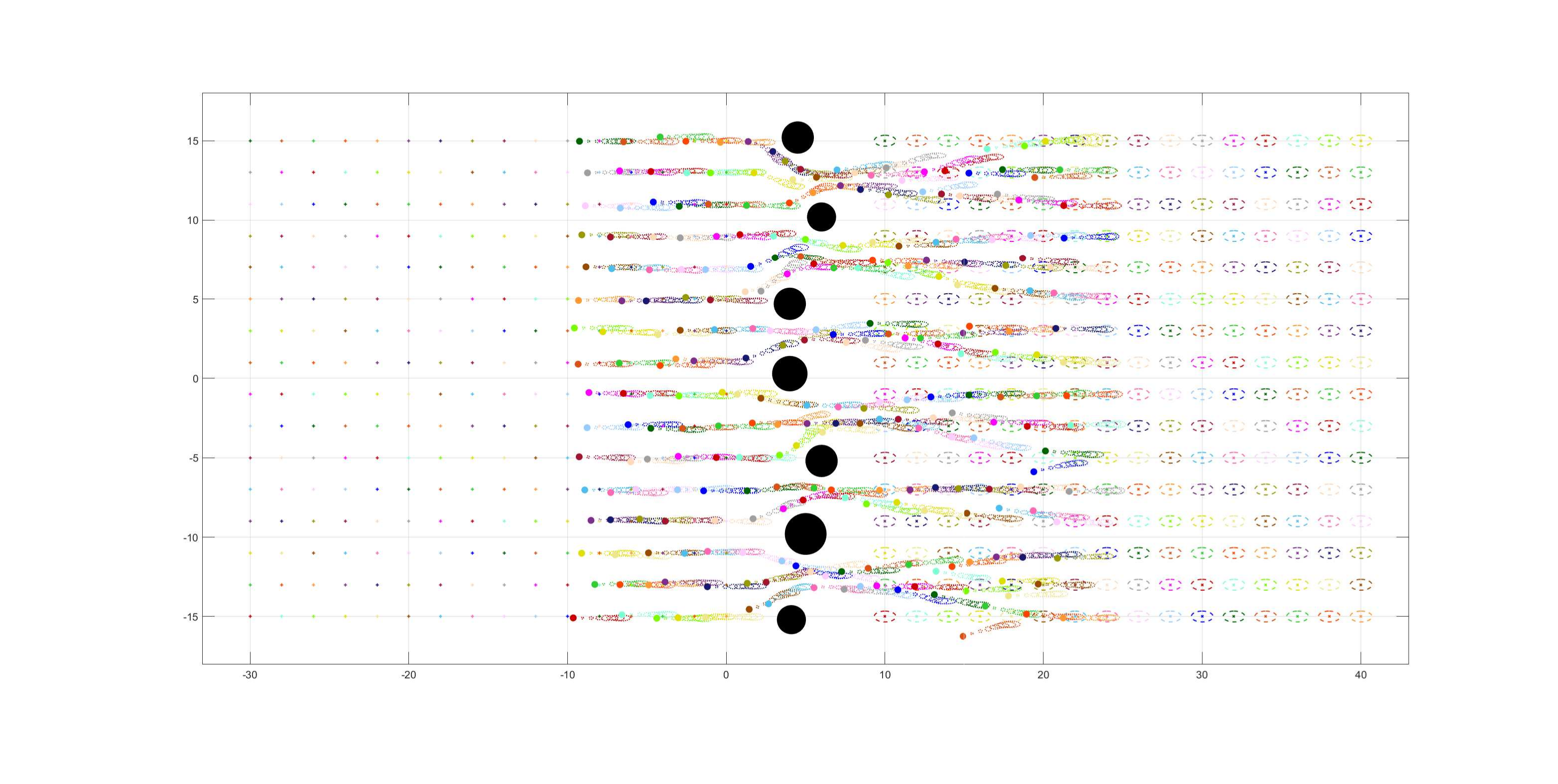}};
\node[align=center, text=NavyBlue] (c) at (8.1, 0.45) {\small{$k = 40$}};
\end{tikzpicture}
\caption{A large-scale task with 256 robots.
}
\label{fig_formation}
\end{figure}

Next, we illustrate the computational and performance advantages of DiMPCS against related SMPC approaches. All comparisons are on the same task as in Fig. \ref{fig_formation}. Initially, we compare against an equivalent centralized approach for solving Problem \ref{multi-robot problem 2}. We observe that as the number of robots grows (Table \ref{tab: comp times}), DiMPCS remains scalable, while the increasing dimensionality of the multi-robot problem makes the centralized approach computationally intractable. We should also highlight that hard-constrained CS approaches that lead to SDPs are excluded from this comparison, as their computational demands are much higher, in addition to their need for a distribution path before performing MPC.

Furthermore, we provide a performance comparison of DiMPCS against standard SMPC methods in terms of collision percentages and control efforts (Table \ref{tab: performance}). Each algorithm is tested for 5 trials. First, we compare against solving the MPC problems with LQG control instead. While the latter method also yields safe solutions, it reduces the variance of the states more aggressively which requires excessive control effort. We also compare with standard SMPC approaches which only optimize for the feed-forward controls, while selecting a fixed stabilizing gain for the initial linearized dynamics \cite{p:arcari2022stochasticMPC}. Although such approaches involve less decision variables, the fact that the covariance is not actively steered leads to either unsafe solutions (Case I) or relatively safe solutions that require significant control effort (Case II). Therefore, the fact that DiMPCS actively steers the state distributions to match target distributions, while computing a sequence of feedback gains, provides the most advantageous combination of safety and control effort.

\section{Hardware Experiments}

Finally, we validate the applicability of the proposed distributed algorithm on a multi-robot system in the Robotarium platform \cite{wilson2020robotarium} at Georgia Tech. For the dynamics of the robots, the reader is referred to \cite{wilson2020robotarium}. In addition to collision and obstacle avoidance constraints with $d_{i,o} = 0.1\text{m}$, $d_{i,j} = 0.2\text{m}$, all robots are subject to the following control constraints, 
%
$
- \bb_{\text{max}} \leq \vG \bu_i  \leq \bb_{\text{max}}, 
$
%
with 
$
\vG = (1/2R) [2, L ; 2 , -L]
$
and
$
\bb_{\text{max}} = [v_{\text{wheel}}^{\text{max}}; v_{\text{wheel}}^{\text{max}}]
$,
%
%
where $R = 0.016$m is the wheel radius, $L = 0.11$m is the axle length and $v_{\text{wheel}}^{\text{max}} = 12.5$ \text{rad/s} is the maximum wheel speed.
The control constraints are handled as chance constraints of the form \eqref{control chance constraint} with $\beta = 0.997$. The timestep is $dt = 330\text{ms}$, while we set $N_{\mathrm{pred}} = 7$ and $N_{\mathrm{total}} = 100$.

\begin{figure*}[!t]
     \centering
     \begin{subfigure}[b]{0.19\textwidth}
         \centering 
         \begin{tikzpicture}
        \node[anchor=south west,inner sep=0] at (0,0){    \includegraphics[width=\textwidth, trim={0cm 0cm 0cm 0cm},clip]{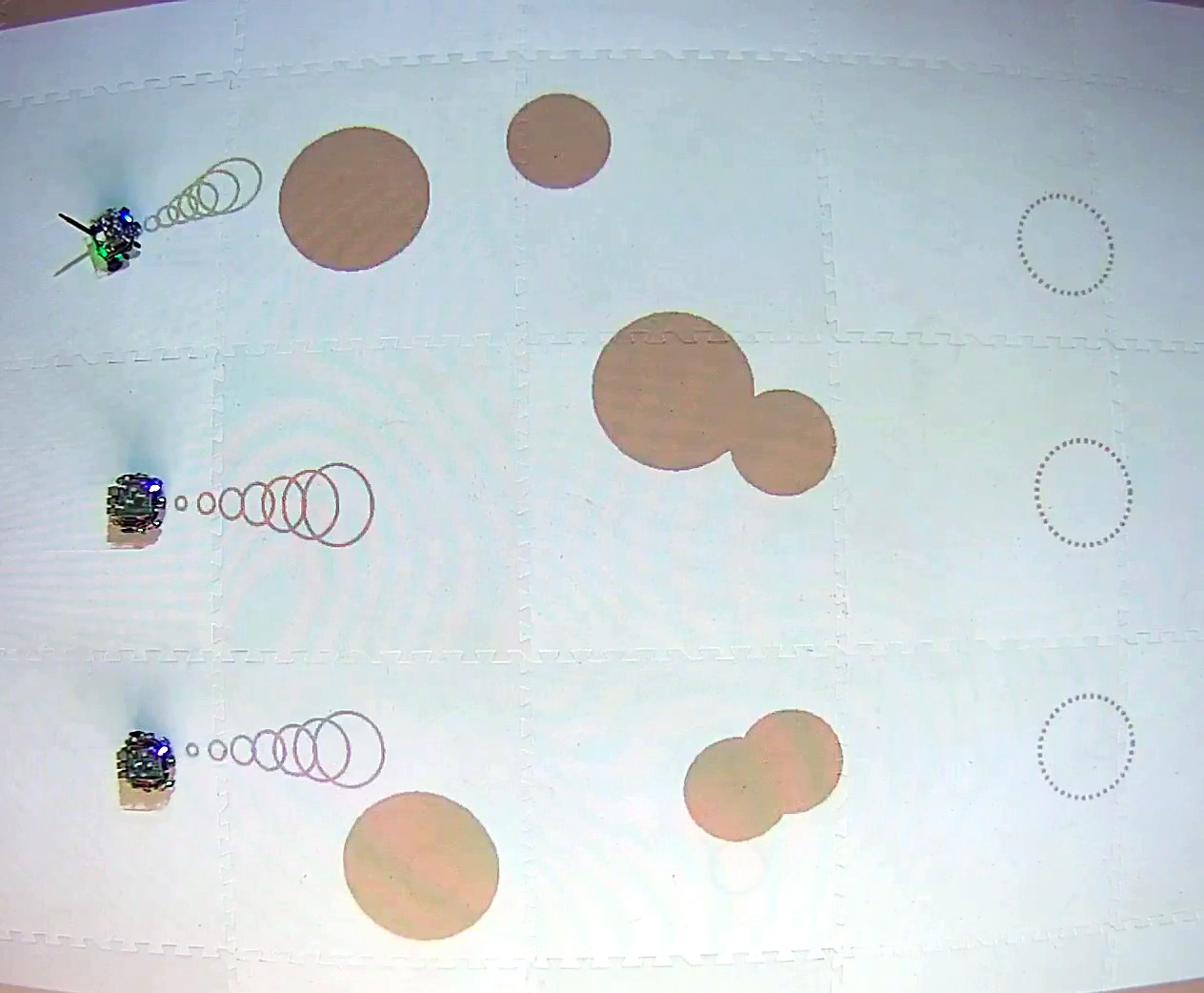}};
        \node[align=center, text=NavyBlue] (c) at (2.88, 0.27) {\small{$t = 0\text{s}$}};
        \end{tikzpicture}
        \label{fig_bottle_4}
    \end{subfigure}
         \begin{subfigure}[b]{0.19\textwidth}
         \centering 
      \begin{tikzpicture}
        \node[anchor=south west,inner sep=0] at (0,0){    \includegraphics[width=\textwidth, trim={0cm 0cm 0cm 0cm},clip]{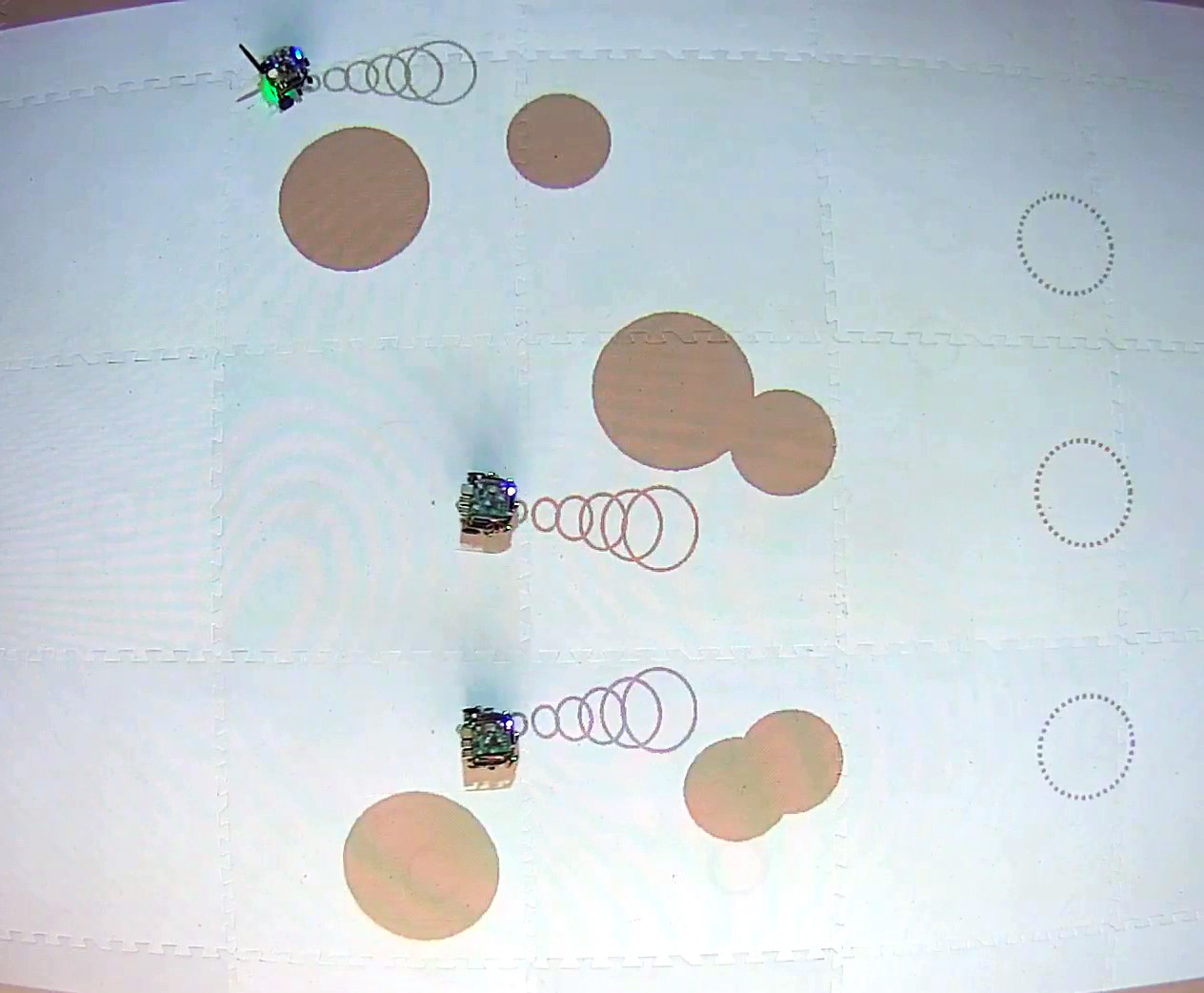}};
        \node[align=center, text=NavyBlue] (c) at (2.9, 0.27) {\small{$t = 5\text{s}$}};
        \end{tikzpicture}
        \label{fig_bottle_4}
    \end{subfigure}
     \begin{subfigure}[b]{0.19\textwidth}
         \centering 
       \begin{tikzpicture}
        \node[anchor=south west,inner sep=0] at (0,0){    \includegraphics[width=\textwidth, trim={0cm 0cm 0cm 0cm},clip]{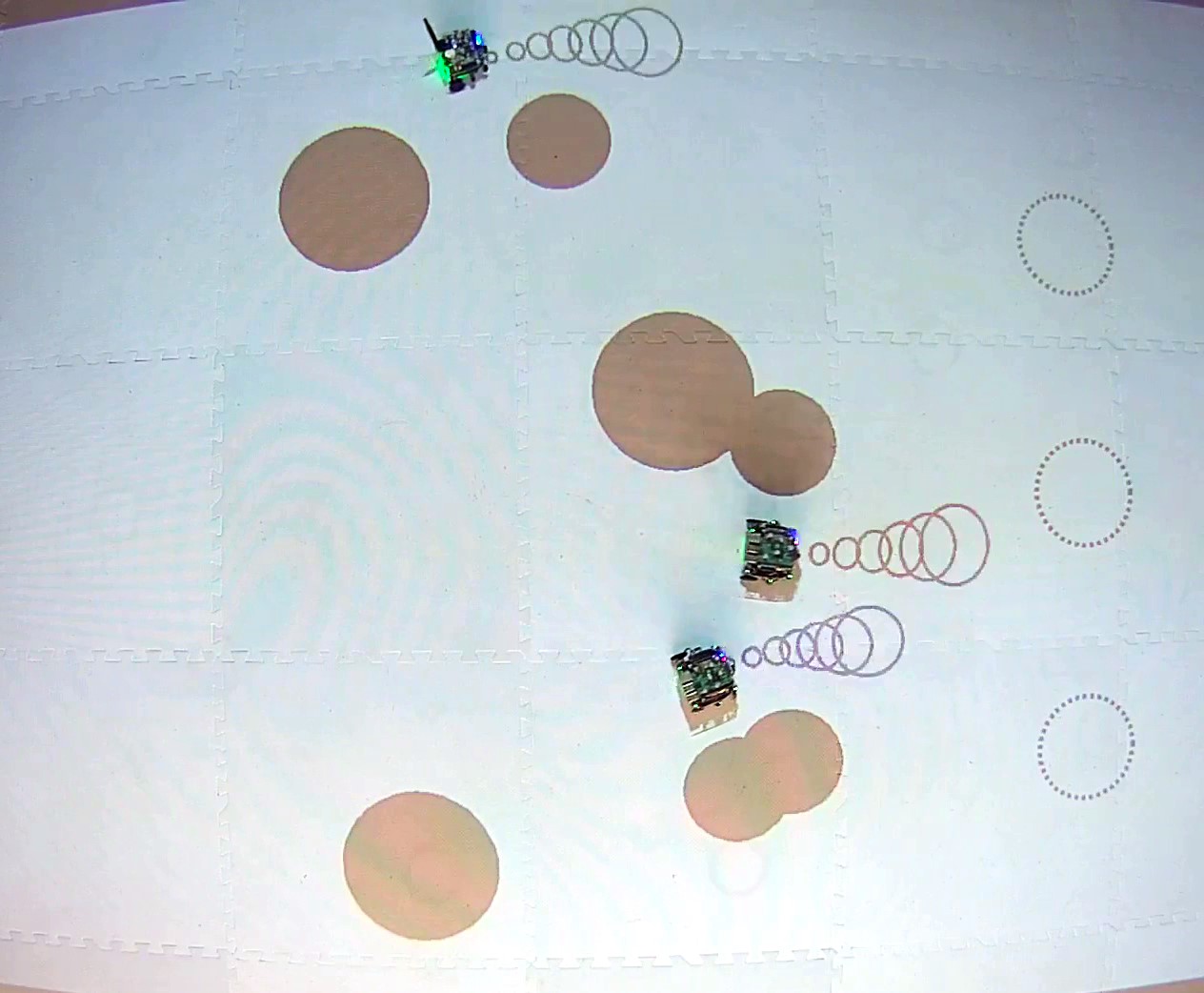}};
        \node[align=center, text=NavyBlue] (c) at (2.9, 0.27) {\small{$t = 8\text{s}$}};
        \end{tikzpicture}
        \label{fig_bottle_4}
    \end{subfigure}
     \centering
     \begin{subfigure}[b]{0.19\textwidth}
         \centering 
       \begin{tikzpicture}
        \node[anchor=south west,inner sep=0] at (0,0){    \includegraphics[width=\textwidth, trim={0cm 0cm 0cm 0cm},clip]{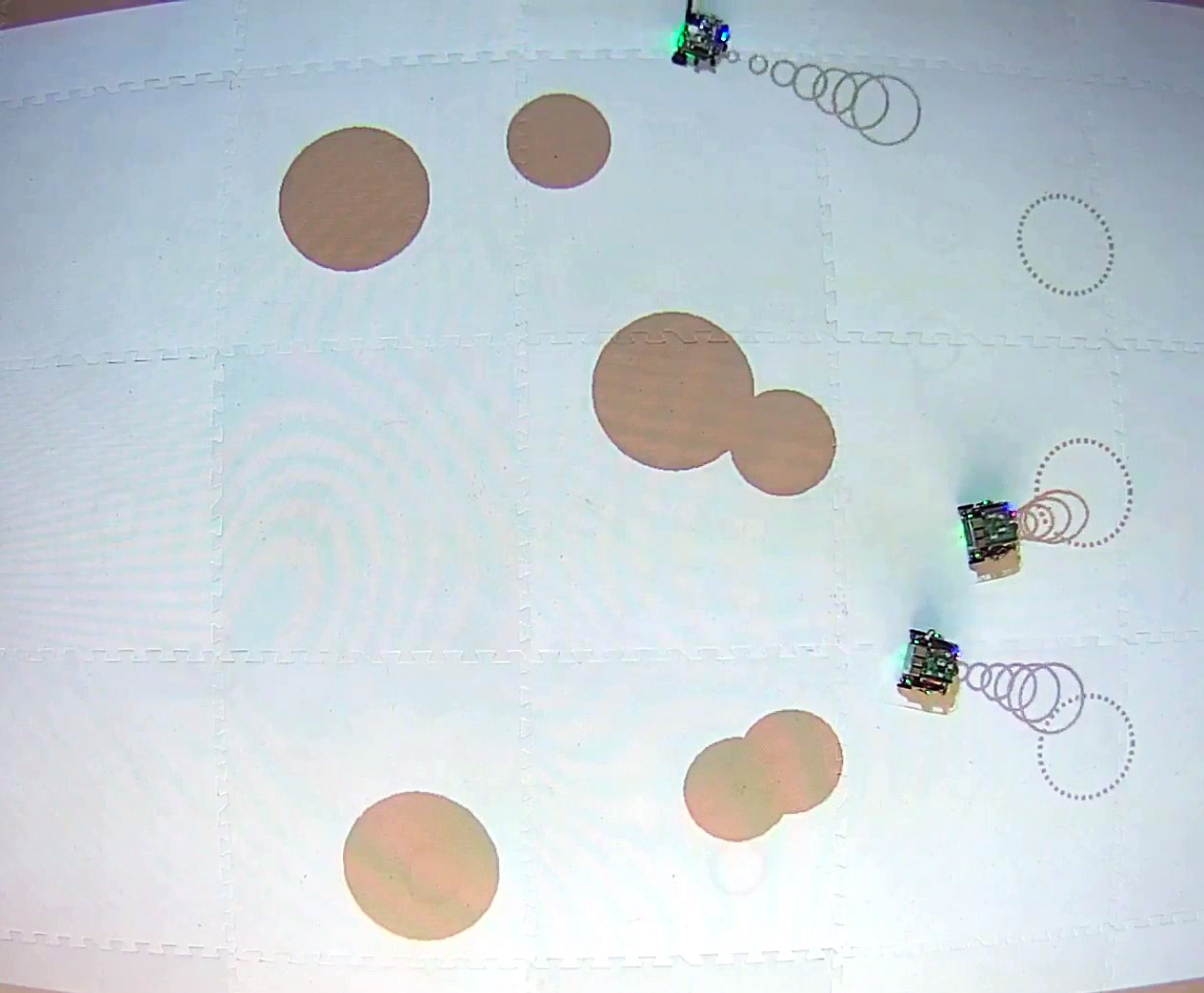}};
        \node[align=center, text=NavyBlue] (c) at (2.8, 0.27) {\small{$t = 11\text{s}$}};
        \end{tikzpicture}
        \label{fig_bottle_4}
    \end{subfigure}
     \begin{subfigure}[b]{0.19\textwidth}
         \centering 
        \begin{tikzpicture}
        \node[anchor=south west,inner sep=0] at (0,0){    \includegraphics[width=\textwidth, trim={0cm 0cm 0cm 0cm},clip]{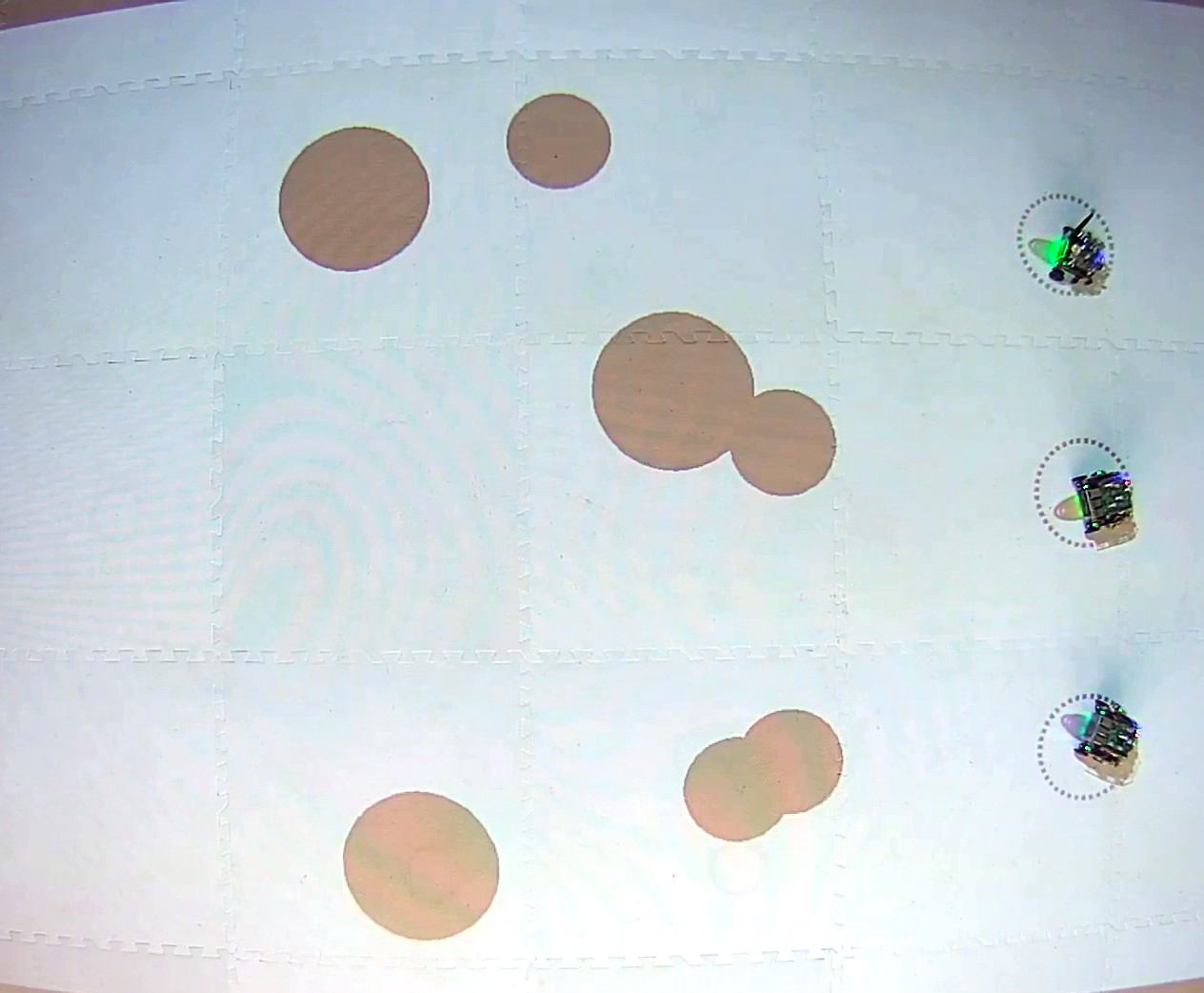}};
        \node[align=center, text=NavyBlue] (c) at (2.8, 0.27) {\small{$t = 20\text{s}$}};
        \end{tikzpicture}
        \label{fig_bottle_4}
    \end{subfigure}
    \hfill
    \vspace{-0.3cm}
\caption{Hardware experiment with three robots that are required to reach their targets while avoiding collisions.}
\label{fig: robotarium_obs_1}
\end{figure*}

\begin{figure*}[!t]
\vspace{-0.1cm}
     \centering
     \begin{subfigure}[b]{0.19\textwidth}
         \centering 
          \begin{tikzpicture}
        \node[anchor=south west,inner sep=0] at (0,0){    \includegraphics[width=\textwidth, trim={0cm 0cm 0cm 0cm},clip]{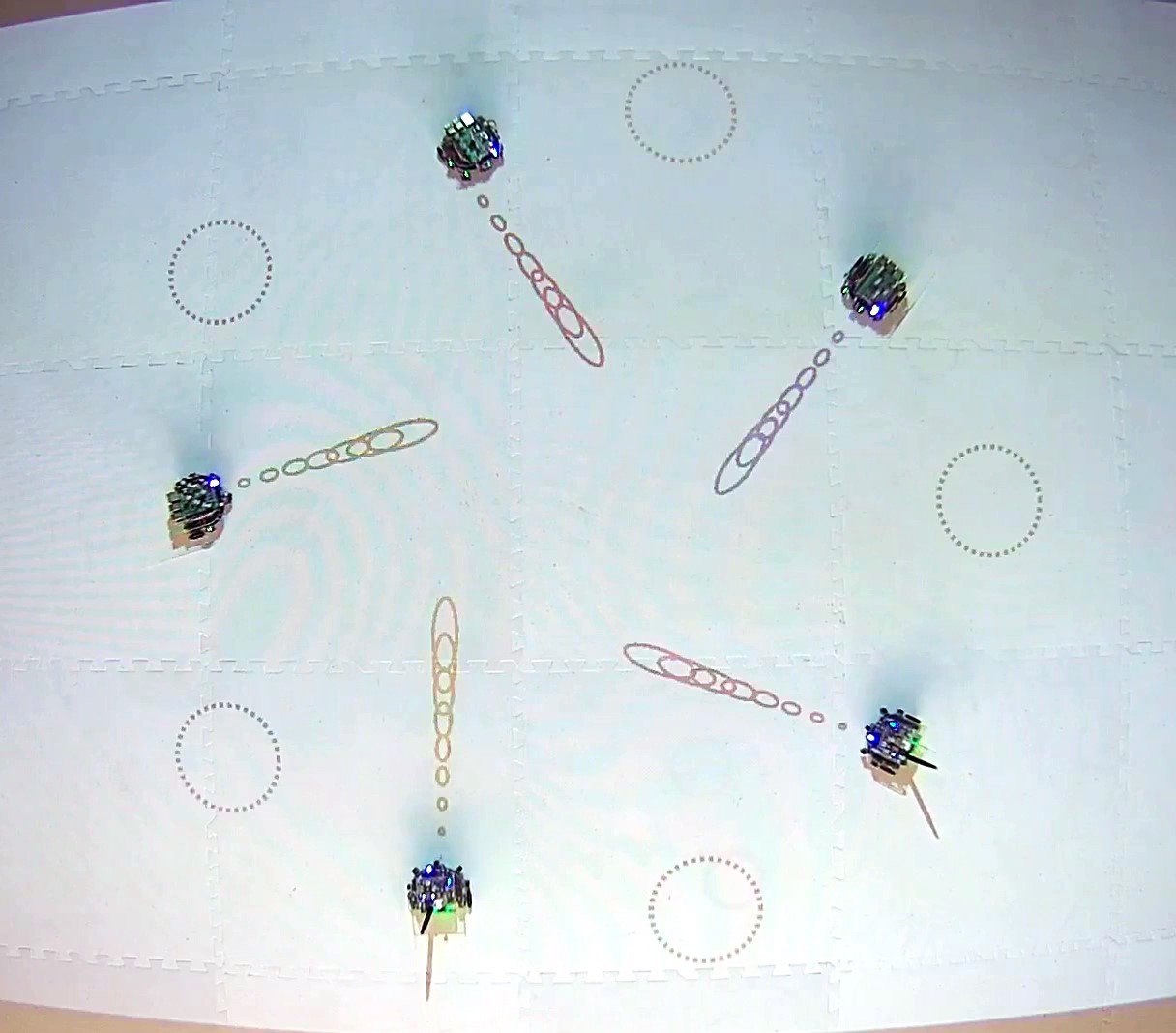}};
        \node[align=center, text=NavyBlue] (c) at (2.88, 0.27) {\small{$t = 0\text{s}$}};
        \end{tikzpicture}
        \label{fig_bottle_4}
    \end{subfigure}
         \begin{subfigure}[b]{0.19\textwidth}
         \centering 
           \begin{tikzpicture}
        \node[anchor=south west,inner sep=0] at (0,0){    \includegraphics[width=\textwidth, trim={0cm 0cm 0cm 0cm},clip]{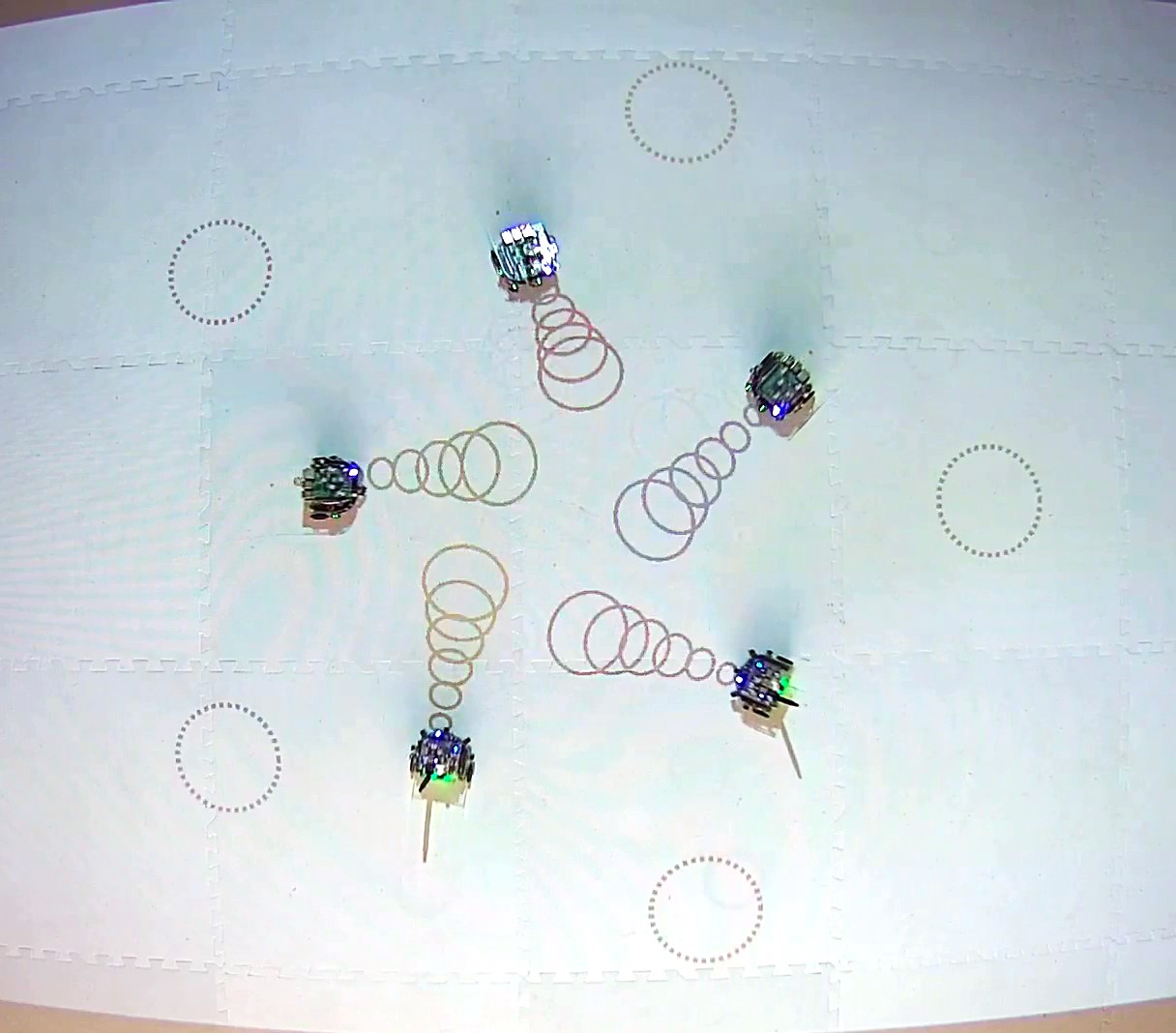}};
        \node[align=center, text=NavyBlue] (c) at (2.88, 0.27) {\small{$t = 2\text{s}$}};
        \end{tikzpicture}
        \label{fig_bottle_4}
    \end{subfigure}
     \begin{subfigure}[b]{0.19\textwidth}
         \centering 
           \begin{tikzpicture}
        \node[anchor=south west,inner sep=0] at (0,0){    \includegraphics[width=\textwidth, trim={0cm 0cm 0cm 0cm},clip]{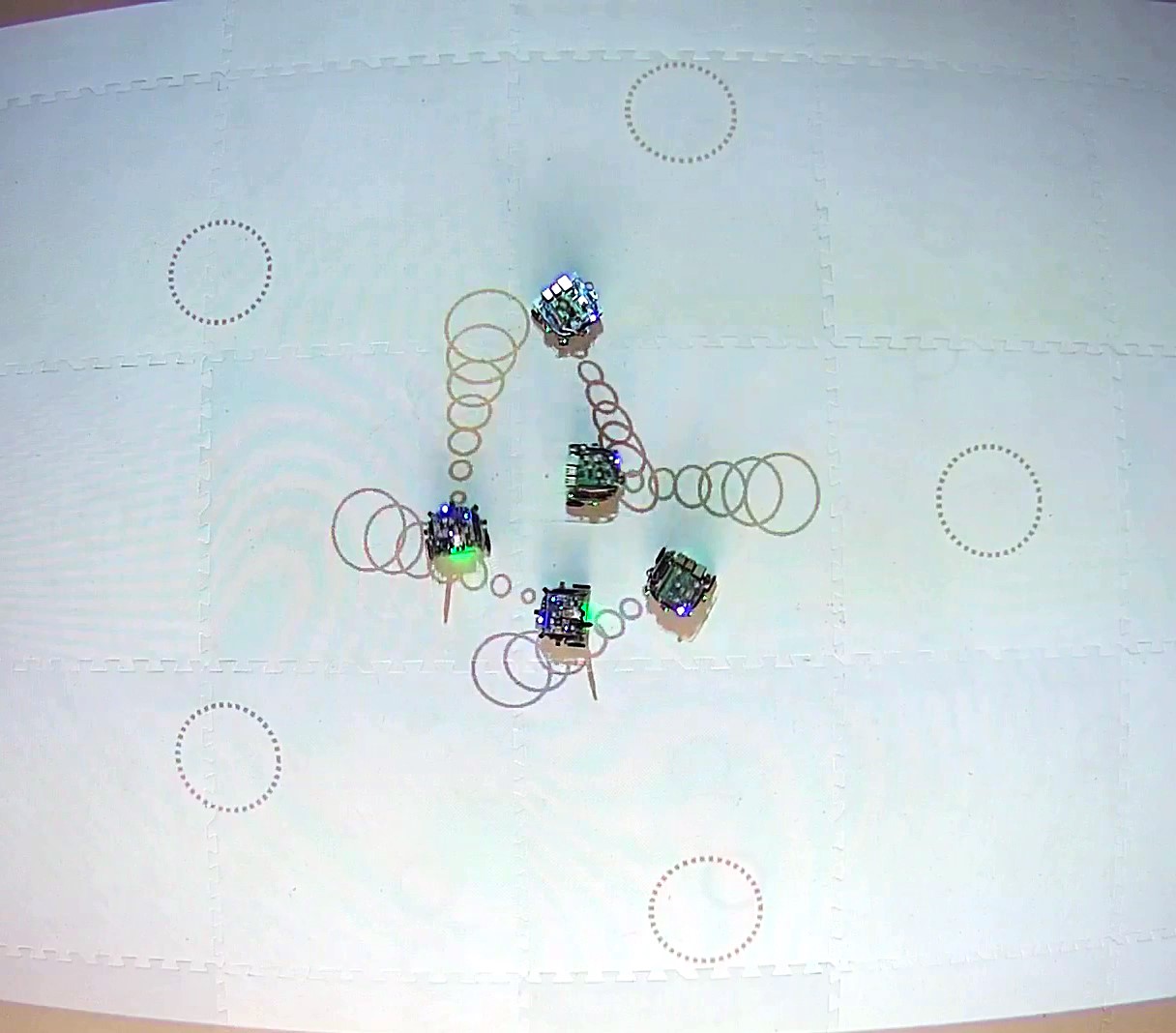}};
        \node[align=center, text=NavyBlue] (c) at (2.88, 0.27) {\small{$t = 4\text{s}$}};
        \end{tikzpicture}
        \label{fig_bottle_4}
    \end{subfigure}
     \centering
     \begin{subfigure}[b]{0.19\textwidth}
         \centering 
           \begin{tikzpicture}
        \node[anchor=south west,inner sep=0] at (0,0){    \includegraphics[width=\textwidth, trim={0cm 0cm 0cm 0cm},clip]{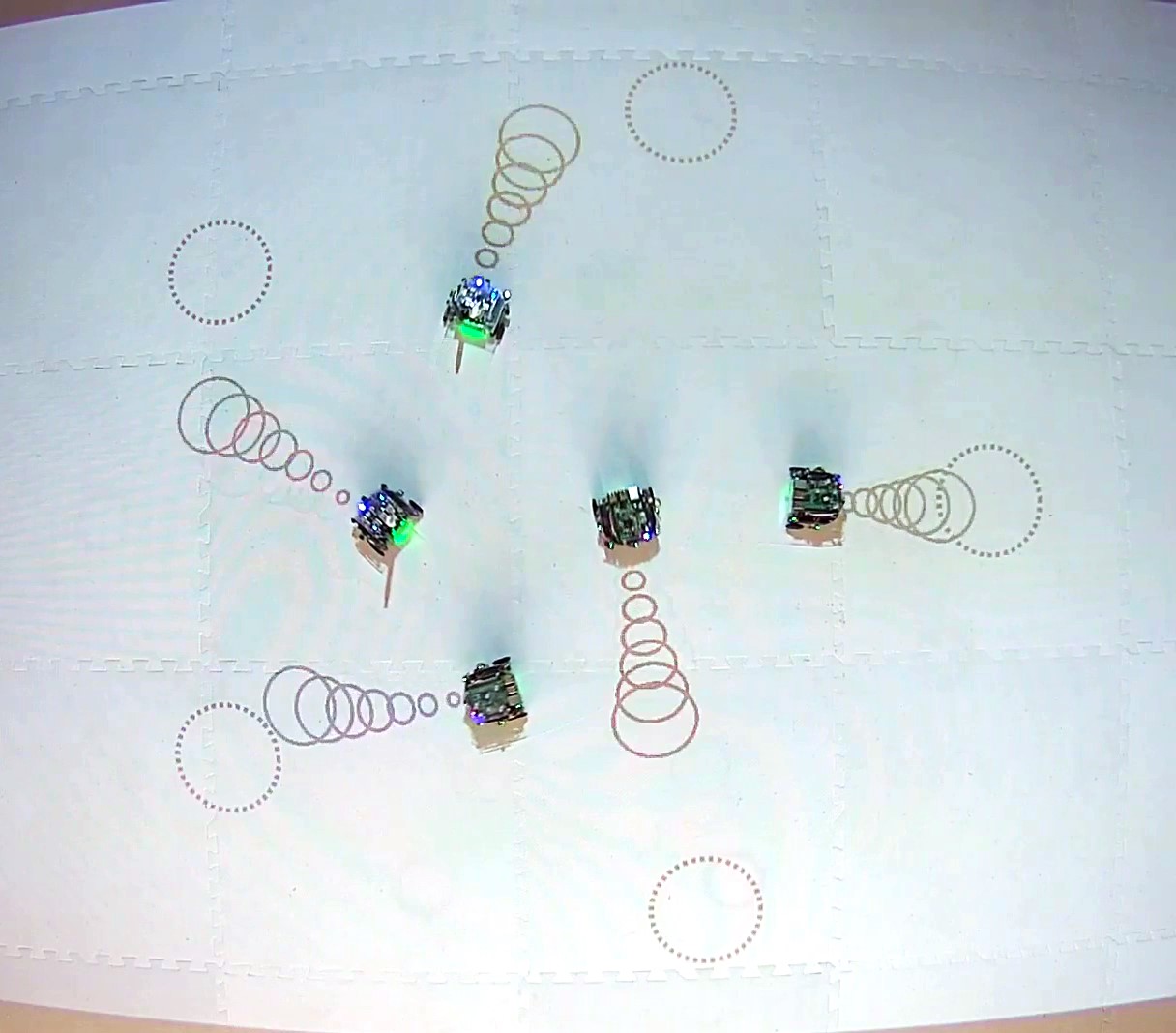}};
        \node[align=center, text=NavyBlue] (c) at (2.88, 0.27) {\small{$t = 7\text{s}$}};
        \end{tikzpicture}
        \label{fig_bottle_4}
    \end{subfigure}
     \begin{subfigure}[b]{0.19\textwidth}
         \centering 
           \begin{tikzpicture}
        \node[anchor=south west,inner sep=0] at (0,0){    \includegraphics[width=\textwidth, trim={0cm 0cm 0cm 0cm},clip]{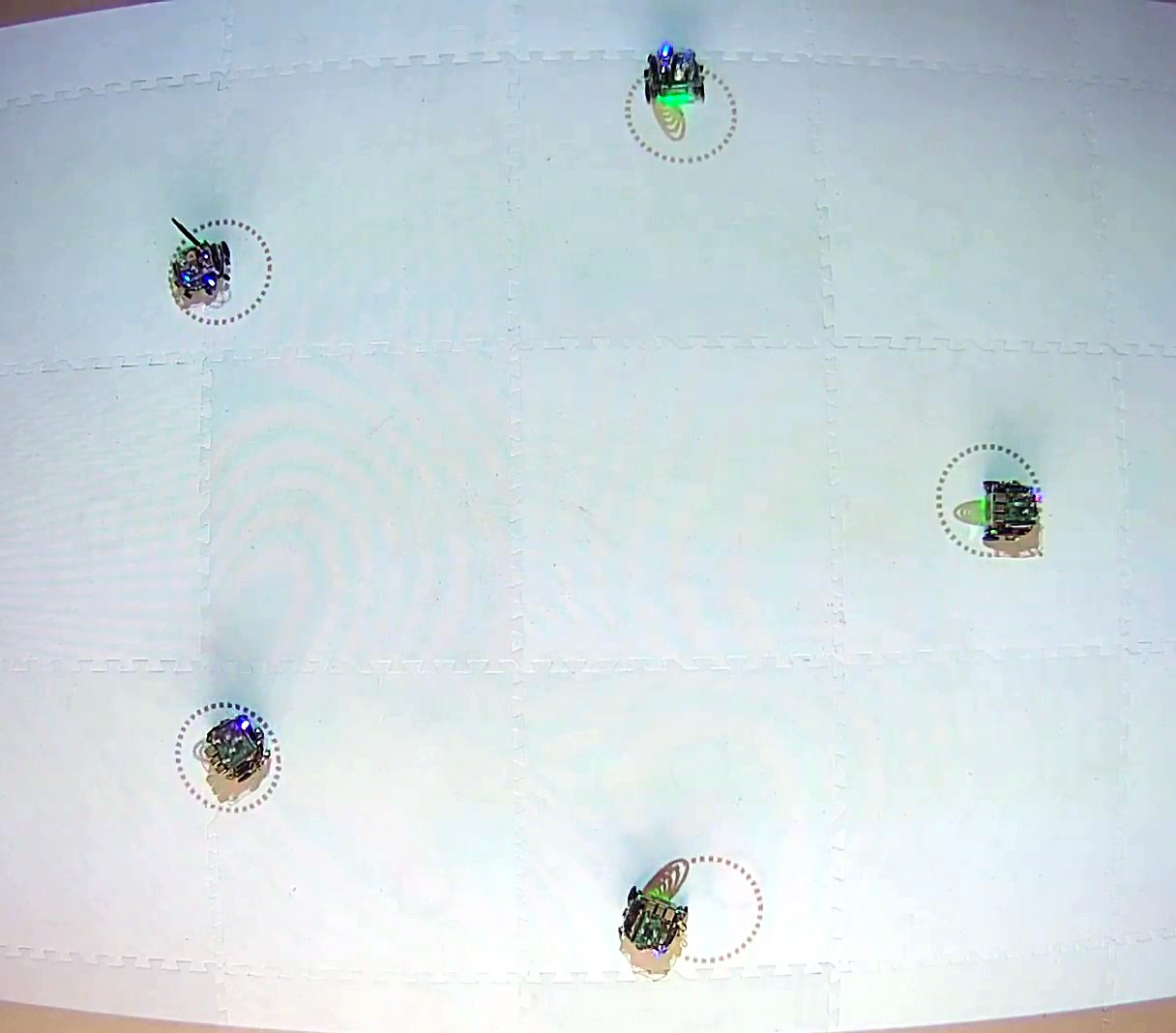}};
        \node[align=center, text=NavyBlue] (c) at (2.78, 0.27) {\small{$t = 20\text{s}$}};
        \end{tikzpicture}
        \label{fig_bottle_4}
    \end{subfigure}
    \hfill
    \vspace{-0.3cm}
\caption{Hardware experiment with five robots that are required to reach the diametrically opposite positions without collisions.}
\label{fig: robotarium_obs_2}
\end{figure*}

We first apply the proposed algorithm on a task where three robots are required to reach to their target distributions while avoiding the obstacles in the middle of the field. As illustrated in Fig. \ref{fig: robotarium_obs_1}, the robots are able to successfully complete the task while avoiding collisions. Next, we demonstrate in Fig. \ref{fig: robotarium_obs_2}, a task where five robots must reach to the diametrically opposite positions while avoiding collisions with the rest of the robots. Again, all robots are safely driven to their destinations without colliding with each other. 

\begin{table}[!t]
\vspace{-0.1cm}
\centering
\begin{tabular}{@{}ccccc@{}}
\hline 
Method & $N=4$ & $N=16$ & $N=64$ & $N=256$ \\
\hline 
\textbf{DiMPCS (Proposed)} & \textbf{321ms} & \textbf{534ms} & \textbf{1.02s} & \textbf{2.05s}  \\
Centralized MPCS & 1.54s & 32s & 9m 49s & 1h 22s  \\
\hline
\end{tabular}
\caption{Computational times per MPC cycle of DiMPCS and an equivalent centralized approach.}
\label{tab: comp times}
\end{table}

\begin{table}[!t]
\centering
\begin{tabular}{@{}ccc@{}}
\hline 
Method & Collisions \% & Control effort  \\
\hline 
\textbf{DiMPCS (Proposed)} & \textbf{0} $\%$ & \textbf{180.55}
\\
SMPC with LQG & 0 $\%$ & 263.83
\\
SMPC with fixed feedback (I) & 5.47 $\%$ & 78.49
\\
SMPC with fixed feedback  (II) & 0.33 $\%$ & 244.73
\\
\hline
\end{tabular}
\caption{Performance comparison between DiMPCS and other SMPC approaches.}
\vspace{-0.2cm}
\label{tab: performance}
\end{table}

\section{Conclusion}\label{sec: conclusion}
In this work, we propose DiMPCS, a novel distributed SMPC algorithm for multi-robot control under uncertainty. 
Our approach combines CS theory using the Wasserstein distance and ADMM into an MPC scheme, to ensure safety while achieving scalability and parallelization.
Numerical simulations verify the effectiveness of DiMPCS in various multi-robot navigation problems compared to other approaches. 
Finally, the applicability of the method on real robotic systems is verified through hardware experiments.


\section*{Appendix}



\subsection{Cost and Constraints Expressions}
\label{cost constraints appendix}

Following a similar derivation as in \cite[Propositions 4,5]{balci2021covariance}, the terms $J_i^{\mathrm{dist}}$ and $J_i^{\mathrm{cont}}$ can be written equivalently as 
\vspace{-0.2cm}
\begin{align*}
& J_i^{\mathrm{dist}} (\bar{\bu}_i, \vL_i, \vK_i) = 
\sum_{k=1}^K
\| \vT_{i,k} \beeta_i(\bar{\bu}_i) - \mu_{i,\mathrm{f}} \|_2^2 
\nonumber
\\
& + \| \bzeta_{i,k}(\vL_i, \vK_i) \|_F^2 + \trace(\Sigma_{i,\mathrm{f}}) 
- 2 \|  \sqrt{\Sigma_{i,\mathrm{f}}} \bzeta_{i,k}(\vL_i, \vK_i) \|_*,
\\[0.1cm]
& J_i^{\mathrm{cont}} (\bar{\bu}_i, \vL_i, \vK_i) = 
\bar{\bu}_i^\T \vR_i \bar{\bu}_i
+ \trace(\vR_i \vL_i \Sigma_{i,0} \vL_i^\T)
\nonumber
\\
& ~~~~~ + \trace(\vR_i \vK_i \vW_i \vK_i^\T),
\end{align*}
%
%
where $\vR_i = \bdiag(R_i, \dots, R_i) \in \Rb^{K m_i \times K m_i}$ and
\vspace{-0.2cm}
\begin{equation*}
\bzeta_{i,k}(\vL_i, \vK_i) = \vT_{i,k}
\begin{bmatrix}
\vG_{i,0} + \vG_{i,u} \vL_i & \vG_{i,w} + \vG_{i,u} \vK_i
\end{bmatrix}.
\end{equation*}
Futhermore, the constraints \eqref{obs avoidance 2 mean}, \eqref{collision avoidance 2 mean} can be written as
\begin{align*}
b_i(\bar{\bu}_i) 
& = d_{i,o} + r_o - \| H_i \vT_{i,k} \beeta_i(\bar{\bu}_i) - p_o \|_2 \leq 0,
\\
c_{i,j}(\bar{\bu}_i, \bar{\bu}_j) 
& = \| H_i \vT_{i,k} \beeta_i(\bar{\bu}_i) - H_j \vT_{j,k} \beeta_j(\bar{\bu}_j) \|_2 \leq 0.
\end{align*}
%

\addtolength{\textheight}{0cm}   

\section*{Acknowledgment}

This work was supported in part by NSF under grants 1936079 and 1937957, and by the ARO Award $\#$W911NF2010151. Augustinos Saravanos acknowledges financial support by the A. Onassis Foundation Scholarship.


\bibliographystyle{IEEEtran}

\bibliography{references}

\end{document}